%% file: arxiv.tex
\documentclass[10pt,twocolumn,letterpaper]{article}
\usepackage[pagenumbers]{cvpr}
\input{preamble}
\definecolor{cvprblue}{rgb}{0.21,0.49,0.74}
\usepackage[breaklinks,colorlinks,allcolors=cvprblue]{hyperref}
\hypersetup{pdftitle={UniSTEM: A Unified Multimodal Model for Electron-Microscopy Understanding and Generation},
  pdfauthor={Linghao Kong, Yunfan Yang, Hanling Zhang, Zixuan Hu, Guanyu Chen, Ben Fei, Shixiang Tang, Xiaopeng Hong, Wangmeng Zuo}}

\title{UniSTEM: A Unified Multimodal Model for Electron-Microscopy Understanding and Generation}
\author{%
Linghao Kong\textsuperscript{1,2}, Yunfan Yang\textsuperscript{3}, Hanling Zhang\textsuperscript{4},
Zixuan Hu\textsuperscript{5}, Guanyu Chen\textsuperscript{3}\\
Ben Fei\textsuperscript{2,4}\thanks{Co-corresponding authors: Ben Fei (\texttt{feynben@gmail.com}) and Shixiang Tang (\texttt{shawn.tang@ox.ac.uk}).},
Shixiang Tang\textsuperscript{2,6}\footnotemark[1],
Xiaopeng Hong\textsuperscript{1}, Wangmeng Zuo\textsuperscript{1}\\[0.5em]
\textsuperscript{1}Harbin Institute of Technology \quad
\textsuperscript{2}Shanghai Artificial Intelligence Laboratory\\
\textsuperscript{3}Fudan University \quad
\textsuperscript{4}The Chinese University of Hong Kong\\
\textsuperscript{5}Peking University \quad
\textsuperscript{6}University of Oxford\\[0.3em]
\texttt{leonardokong486@gmail.com}
}

\begin{document}
\maketitle
\input{sec/0_abstract}
\input{sec/1_intro}
\input{sec/2_related_work}

\input{sec/3_task_benchmark_arxiv}
\input{sec/4_method_arxiv}
\input{sec/5_experiments_arxiv}
\input{sec/conclusion}

\clearpage
{\small
\bibliographystyle{ieeenat_fullname_citeorder}
\bibliography{main}
}

\clearpage
\onecolumn
\suppressfloats[t]
\renewcommand{\floatpagefraction}{0.8}
\renewcommand{\topfraction}{0.95}
\renewcommand{\textfraction}{0.05}
\section*{Appendix}
\input{sec/6_appendix_arxiv}
\end{document}

%% file: preamble.tex
\usepackage{multirow}
\newcommand{\singlelinehead}[1]{\raisebox{.5\normalbaselineskip}{#1}}
\usepackage[ruled]{algorithm}
\usepackage{algpseudocode}



%% file: sec/0_abstract.tex
\begin{abstract}
In electron microscopy for materials research, general-purpose multimodal models struggle with atomic-scale STEM tasks, while many specialized models address only isolated tasks or narrow workflows. We introduce a six-task benchmark spanning understanding and generation across common STEM materials-analysis needs, including defect detection, crystal-structure and conductivity prediction, differential phase contrast (DPC) field reconstruction, and denoising. Different output objectives can cause interference under joint training, while uneven task performance calls for targeted feedback. To address these challenges, we propose \textbf{UniSTEM}, a unified model built through weak-task identification and feedback, isolated training of understanding and generation modules, and assembly of parallel branches. UniSTEM outperforms task-specific experts on 9 of 13 reported metrics, improving dislocation-detection F1 by 6.7\% and reducing conductivity log-MAE by 15.8\%. Gains across tasks and imaging conditions suggest that these related capabilities can develop together and support the effectiveness of our multitask training approach for electron microscopy.
\end{abstract}

%% file: sec/1_intro.tex
\section{Introduction}
\label{sec:intro}

Atomic-resolution scanning transmission electron microscopy (STEM) reveals crystal structures and defects at the scale of individual atomic columns~\cite{leitherer2023automatic}. Advances in differential phase contrast (DPC) and 4D-STEM further enable electric-field imaging at atomic resolution~\cite{shibata2012differential,hachtel2018sub}. Analysis at this scale requires distinguishing nearby atomic sites and preserving fine spatial patterns in reconstructed fields and images. Existing methods have identified structures and interfaces in atomic-resolution STEM~\cite{leitherer2023automatic}, while vision-language models have enabled interpretation across diverse TEM images~\cite{tu2026atomic}.

These advances address different parts of the analysis workflow. Specialized methods focus on particular predictions, and integrated systems connect multiple steps through separate components~\cite{chen2026bridging}. Whether one model can retain specialist performance across atomic-site localization, structured material predictions, and spatially faithful image outputs remains unclear. These outputs require different supervision and training objectives, and multitask optimization can degrade individual capabilities~\cite{yu2020gradient}. A general-purpose STEM model therefore needs evaluation across these tasks and a training strategy that reinforces capabilities lagging behind task experts.

We address the evaluation gap with a benchmark of six STEM tasks comprising four understanding tasks and two generation tasks. They cover defect detection and crystal-structure characterization~\cite{chu2024accurate,leitherer2023automatic}, conductivity prediction, DPC field reconstruction~\cite{shibata2012differential}, and image denoising~\cite{lobato2024deep}, with outputs ranging from structured predictions to reconstructed fields and images. Built from shared crystal-structure and transport-property records~\cite{ricci2017transport} within a common STEM simulation framework, the benchmark connects stages of materials analysis relevant to integrated systems~\cite{chen2026bridging}. Structure-disjoint splits and task-specific metrics enable comparisons of specialized models and general-purpose models across imaging conditions.

To address the training challenge, we propose \textbf{UniSTEM}, a unified multimodal model adapted from the pretrained BAGEL-7B-MoT backbone~\cite{deng2025bagel}. We initialize the model from a selected task expert to provide a strong domain starting point. We then train its understanding and generation towers in sequence, using interim evaluation and expert feedback to reinforce tasks that lag behind their experts. Finally, we refine the towers in separate parallel branches and assemble their modules into one model.

Our contributions are summarized as follows:
\begin{itemize}
    \item We introduce a six-task STEM benchmark spanning understanding and image generation, with structure-disjoint evaluation across imaging conditions.
    \item We develop UniSTEM, a general-purpose model for these tasks, together with an expert-guided training strategy based on targeted feedback, separate tower optimization, and parallel model assembly.
    \item UniSTEM exceeds task experts on 9 of 13 metrics. Ablations support the design, and serial inference probes downstream error propagation.
\end{itemize}

%% file: sec/2_related_work.tex
\section{Related work}
\label{sec:related}

Scanning transmission electron microscopy (STEM) plays a central role in materials characterization by resolving atomic arrangements, defects, and interfaces. Differential phase-contrast STEM further provides access to local electromagnetic fields at atomic resolution~\cite{shibata2012differential}. Interpreting STEM data involves defect localization, crystallographic analysis, image restoration, and material-property inference. Multislice simulation methods provide controlled images~\cite{ophus2017fast,pryor2017streaming}, while transport databases supply material-level property records~\cite{ricci2017transport}. These tasks span atomic sites, crystal structures, material properties, and images, requiring different predictions and evaluations.

Specialized learning-based models have advanced individual steps of electron-microscopy analysis. AI-STEM identifies crystal structures, orientations, and interfaces~\cite{leitherer2023automatic}. Defect-focused methods detect lattice anomalies, vacancies, and low-contrast atoms~\cite{guo2021defect,chu2024accurate}. AtomSegNet addresses atomic-column segmentation and localization alongside image enhancement~\cite{lin2021atomsegnet}. For restoration, Noise2Atom and convolutional models denoise electron micrographs~\cite{wang2020noise2atom,lobato2024deep}, while a UNIT-based model learns STEM denoising from unpaired experimental and simulated images~\cite{wu2026unsupervised}. STEMDiff instead generates STEM images from structural labels~\cite{bao2025stemdiff}. These methods exploit domain-specific image structure, but are generally designed around particular input-output mappings.

Multimodal models have begun to connect electron micrographs and materials data with language. ATOMIC supports TEM question answering~\cite{tu2026atomic}, and MicroscopyGPT predicts atomic structures from simulated STEM images~\cite{choudhary2025microscopygpt}. MatterChat combines crystal structures and text for materials analysis~\cite{tang2026matterchat}, while EMSeek coordinates specialized microscopy and materials-analysis tools~\cite{chen2026bridging}. MatCha evaluates materials-characterization image questions~\cite{lai2025matcha}, and EMVista assesses microstructural understanding in electron micrographs~\cite{wen2026innovator}. Together, these works broaden the scope of scientific image interpretation and its evaluation.

Although general-purpose multimodal models have advanced rapidly, from understanding-oriented Qwen3-VL and InternVL3.5 to BAGEL and SenseNova-U1.5, which span understanding and generation~\cite{bai2025qwen3,wang2025internvl3_5,deng2025bagel,sensenova2026sensenovau1.5}, their capabilities in materials science have not been systematically validated. Researchers have also adapted pretrained foundations to domain tasks. BiomedGPT covers biomedical vision, language, and vision-language tasks~\cite{zhang2024generalist}, Uni-Med uses a mixture-of-experts connector to reduce competition among medical tasks~\cite{zhu2024uni}, and S1-Omni-Image supports scientific image understanding, generation, and editing~\cite{li2026s1omniimage}. Our work applies this generalist approach to materials electron microscopy while seeking to retain performance across tasks.

%% file: sec/3_task_benchmark_arxiv.tex
\section{Tasks and benchmark}
\label{sec:task_benchmark}

We specify the benchmark tasks and describe their construction and data splits below.

\subsection{Tasks}
\label{sec:task_overview}
\label{sec:task_formulation}

\begin{figure}[t]
  \centering
  \includegraphics[width=\columnwidth]{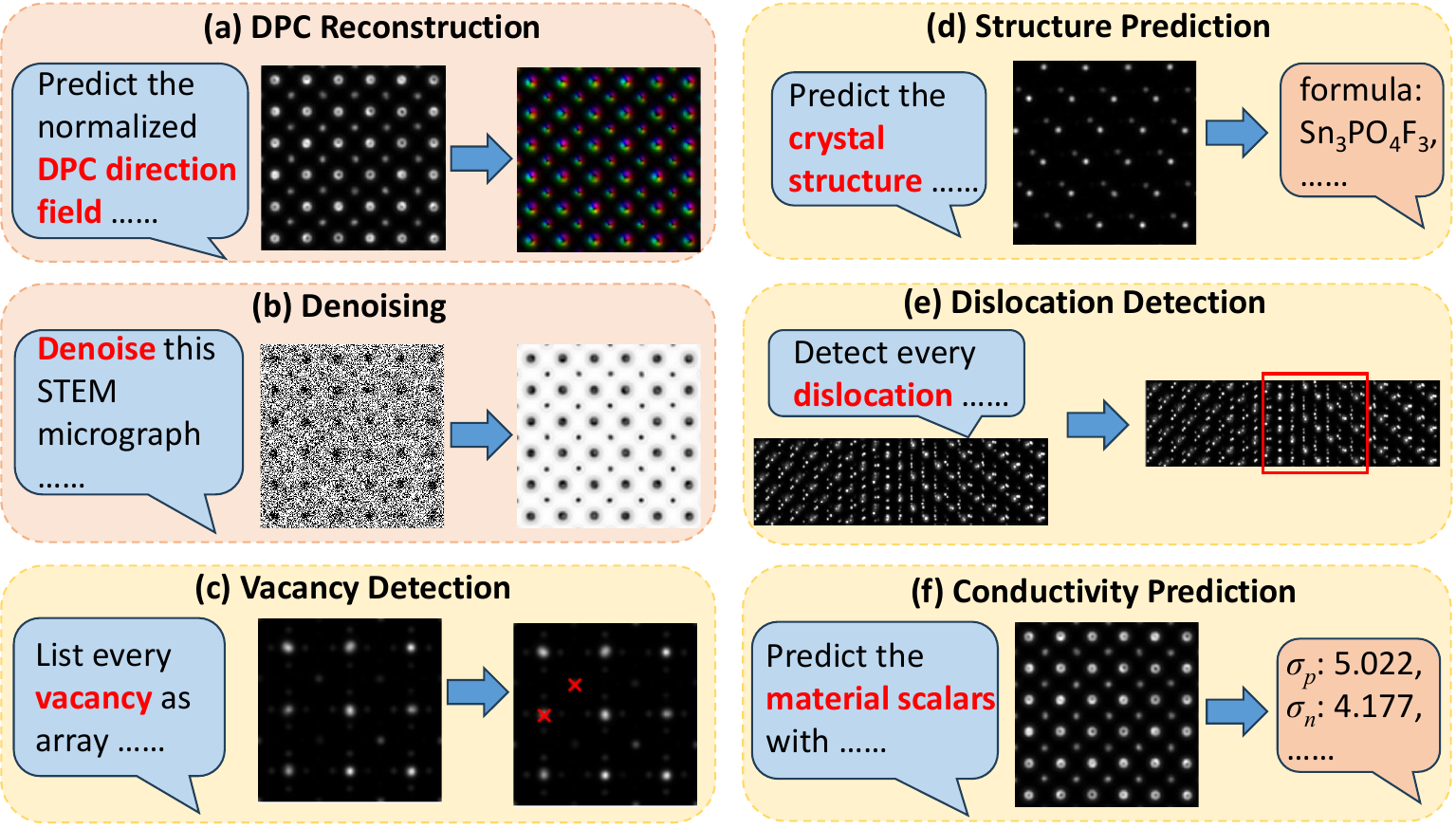}
  \caption{Six STEM tasks spanning structured prediction (U1--U4) and image generation (G1--G2).}
  \label{fig:task_overview}
\end{figure}

The benchmark comprises four structured-prediction tasks (U1--U4) and two image-generation tasks (G1--G2). \Cref{fig:task_overview} illustrates the tasks, and \cref{tab:tasks} specifies their visual inputs, additional context, and prediction targets.

\begin{table*}[t]
  \centering
  \caption{Six benchmark tasks grouped by output modality. Imaging settings comprise the available acquisition conditions. Constituent elements are supplied in the prompts for U2, U4, and G1.}
  \label{tab:tasks}
  \small
  \setlength{\tabcolsep}{3.5pt}
  \begin{tabular}{@{}llllll@{}}
    \toprule
    \textbf{Family} & \textbf{ID} & \textbf{Task} & \textbf{Visual input} & \textbf{Additional context} & \textbf{Prediction target} \\
    \midrule
    \multirow{4}{*}{\textbf{Understanding}} & U1 & Vacancy detection & STEM & Imaging settings & Vacancy coordinates (JSON) \\
     & U2 & Structure prediction & STEM & Settings + element set & Formula, lattice, orientation (JSON) \\
     & U3 & Dislocation detection & STEM & Imaging settings & Dislocation boxes (JSON) \\
     & U4 & Conductivity prediction & STEM/DPC & Settings + element set & $\sigma_p,\sigma_n$ (JSON) \\
    \midrule
    \multirow{2}{*}{\textbf{Generation}} & G1 & DPC field reconstruction & STEM & Settings + element set & Direction map + scale \\
     & G2 & Denoising & Noisy STEM & Denoising instruction & Clean STEM image \\
    \bottomrule
  \end{tabular}
\end{table*}

U1 localizes missing atomic sites, U3 detects dislocations, and U2 predicts chemical formula, lattice parameters, and zone-axis orientation given the constituent element set. U4 predicts $n$- and $p$-type conductivity. Their targets are serialized as JSON. G1 reconstructs a differential phase contrast (DPC) direction-map image and its normalization scale. G2 restores a clean micrograph from a Poisson-noisy input at signal-to-noise ratios (SNRs) of $2$, $5$, $10$, and $20$, conditioned on a text denoising instruction. G2 training records also include an instruction specifying detector mode and SNR. The available imaging conditions, including detector mode, are supplied with the task prompt for the other tasks. Constituent elements are also supplied in the U4 and G1 prompts. U2's known element set and G1's map-plus-scale representation define the scope of those predictions.

\subsection{Benchmark construction and splits}
\label{sec:data_construction}
\label{sec:dataset_scale}

Crystal structures and transport labels come from a public first-principles database~\cite{ricci2017transport}. We simulate STEM images and DPC signals with PRISM multislice calculations~\cite{ophus2017fast,pryor2017streaming}. Vacancy structures are relaxed using a pretrained interatomic potential~\cite{deng2023chgnet}. Simulations vary detector mode (BF, ABF, ADF, HAADF), accelerating voltage ($200$/$300$\,kV), convergence semi-angle ($21$/$30$\,mrad), defocus, and zone axis. This breadth of tasks and imaging conditions lets us assess both multitask capability and performance across imaging settings. The setup provides exact synthetic labels. Additional simulation details are given in the supplementary material.

The processed sample counts are listed in \cref{sec:supp_benchmark_size}. G1 and U4 share a corpus, while the other tasks have their own sample counts. Training and test sets are disjoint by crystal-structure identifier, so views or noise realizations of one structure do not cross the split. For each task, we randomly reserve 1\% of the training pool as a validation set, excluding these samples from both training and testing.

%% file: sec/4_method_arxiv.tex
\section{Method}
\label{sec:method}

We build \textbf{UniSTEM} by adapting a pretrained dual-tower foundation model through serial tower training, followed by parallel tower completion and assembly. Task-expert preparation and warm-start selection precede these stages.

\subsection{Model and training overview}
\label{sec:architecture}

As shown in \cref{fig:task_overview}, our benchmark requires both discrete structured-text outputs and spatially continuous image outputs, which involve different supervision signals and generation processes. We therefore adopt the pretrained \textbf{BAGEL-7B-MoT} backbone~\cite{deng2025bagel}. It provides separate understanding and generation parameter paths above shared multimodal attention, giving the related STEM tasks a common model and conditioning interface while allowing their output modalities to be optimized separately. Our focus is how to integrate these capabilities through training without redesigning the backbone.

UniSTEM is trained in two stages (\cref{fig:training_pipeline}). \textbf{Stage~1} trains the understanding and generation towers in sequence. \textbf{Stage~2} trains two branches from the same checkpoint and assembles the final model. The trainable and frozen modules during isolated understanding- and generation-tower training are detailed in \cref{sec:supp_tower_paths} and illustrated in \cref{fig:dual_tower_sep_train}. During understanding-task training, the ViT, connector, text embeddings, understanding tower, and language head are updated, while the generation tower is frozen or omitted. During generation-task training, only the generation tower and latent projection modules are updated. The frozen understanding tower and its input encoders still participate in the forward pass to encode text prompts and image conditions for generation. Thus, freezing a tower prevents parameter updates but does not necessarily remove it from computation.

Before these stages, we train five experts using supervised fine-tuning (SFT) alone. Four are single-task experts, and one is a dual-output expert shared by U4 conductivity prediction and G1 DPC field reconstruction. U4 and G1 use the same STEM/DPC corpus, allowing paired supervision for their shared expert. For understanding outputs, SFT is next-token cross-entropy over the assistant tokens of a target JSON sequence $y$ conditioned on image and text input $x$,
\begin{equation}
  \label{eq:sft-und}
  \mathcal{L}_{\mathrm{SFT}}^{\mathrm{und}} = -\sum_{t} \log p_\theta\bigl(y_t \mid x,\, y_{<t}\bigr).
\end{equation}
For generation outputs, SFT is flow-matching mean squared error between the predicted velocity and the ground-truth velocity at a state $x_t$ on the ground-truth flow path,
\begin{equation}
  \label{eq:sft-gen}
  \mathcal{L}_{\mathrm{SFT}}^{\mathrm{gen}} = \mathbb{E}\,\bigl\| v_\theta(x_t, t) - v_t^{\mathrm{GT}} \bigr\|_2^2.
\end{equation}
The experts are then frozen for later distillation. Based on validation-set results, we choose U1 vacancy detection as the warm-start capability and initialize UniSTEM from its expert checkpoint, denoted S0.

\subsection{Serial dual-tower training}
\label{sec:training_strategy}

\begin{figure*}[t]
  \centering
  \includegraphics[width=\textwidth]{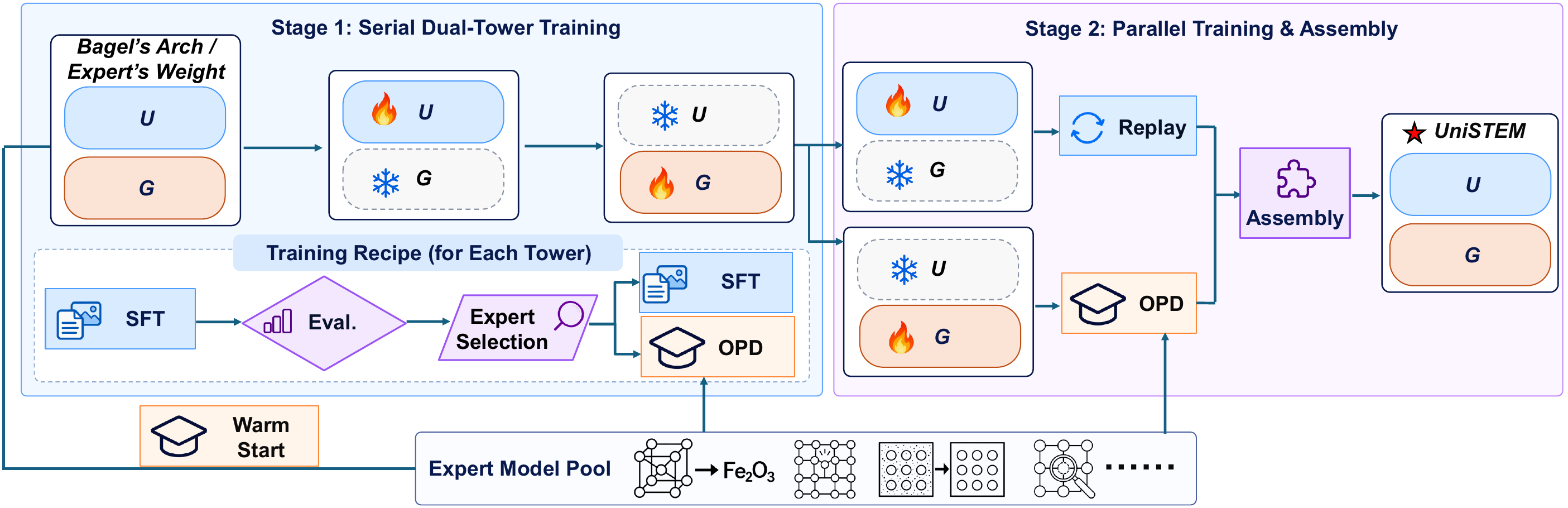}
  \caption{UniSTEM training through expert-guided preparation, Stage~1 serial dual-tower training, and Stage~2 parallel completion and assembly.}
  \label{fig:training_pipeline}
\end{figure*}

The forward dependency between the towers is asymmetric. Generation reads text prompts and available image conditions encoded by the understanding path, while understanding tasks do not require the generation tower. We therefore adapt the understanding path first and then fix it during generation training (\cref{fig:training_pipeline}). This gives generation a stable conditioning representation already adapted to STEM tasks. The architectural dependency motivates this order without assuming that the reverse order cannot work, which we examine separately in \cref{sec:stage_dynamics}.

Mixed supervised fine-tuning first establishes coverage across a task family, but its tasks need not improve at the same rate. We compare the student with task experts on the validation sets $\mathcal D_{\rm val}$ using selected primary metrics and identify tasks $\mathcal W_k$ that lag their experts. Mixed SFT continues on the full task family, with frozen-expert guidance added only on batches from $\mathcal W_k$ (\cref{fig:training_pipeline,alg:training_algorithm,tab:training_stages}).

\begin{algorithm}[t]
  \centering
  \caption{UniSTEM training with expert preparation}
  \label{alg:training_algorithm}
  \footnotesize
  \begin{minipage}{\columnwidth}
  \begin{algorithmic}[1]
    \Statex \textbf{Input:} $\theta_0$, task sets $\mathcal T_{\rm u},\mathcal T_{\rm g}$, validation sets $\mathcal D_{\rm val}$
    \Statex \textbf{Preparation}
    \State $\{E_t\}\gets\operatorname{ExpertTrain}(\theta_0)$
    \State $\theta\gets\operatorname{SelectWarmStart}(\{E_t\},\mathcal D_{\rm val})$
    \Statex \textbf{Stage 1: serial towers}
    \For{$k\in(\mathrm{u},\mathrm{g})$}
      \State \textbf{if} $k=\mathrm g$: $\theta_{\rm g}\gets(\theta_0)_{\rm g}$
      \State $\theta_k\gets\operatorname{SFT}_k(\theta_k)$ \Comment{other tower frozen}
      \State $\mathcal W_k\gets\operatorname{Lagging}(\theta,\{E_t\},\mathcal D_{\rm val})$
      \State $\theta_k\gets\operatorname{SFT{+}OPD}_k(\theta_k;\mathcal W_k)$
    \EndFor
    \Statex \textbf{Stage 2: parallel branches}
    \State $\theta^A,\theta^B\gets\operatorname{Fork}(\theta)$
    \State $\theta^A_{\rm u}\gets\operatorname{SFT}_{\rm u}(\theta^A_{\rm u})$
    \State $\theta^B_{\rm g}\gets\operatorname{SFT{+}OPD}_{\rm g}(\theta^B_{\rm g})$
    \State \textbf{return} $\operatorname{Merge}(\theta^A_{\rm u},\theta^B_{\rm g},\theta^A_{\rm shared})$
  \end{algorithmic}
  \end{minipage}
\end{algorithm}

Starting from S0, we first perform mixed SFT on U1--U4, initially up-sampling U1 (S1-1). Validation-set comparison identifies U2 for expert guidance. We then continue cross-entropy on all four tasks, adding reverse-KL distillation only to U2 batches (S1-2). This loss uses ground-truth answer prefixes rather than student-generated trajectories,
\begin{equation}
  \label{eq:opd-und}
  \mathcal{L}_{\mathrm{OPD}}^{\mathrm{und}} = \sum_{t \in \mathcal{T}_{\mathrm{distill}}} \mathrm{KL}\bigl( p_\theta(\cdot \mid x, y_{<t}) \,\big\|\, q_t(\cdot) \bigr).
\end{equation}
Here $q_t$ is the frozen expert's next-token distribution and $\mathcal{T}_{\mathrm{distill}}$ contains assistant-token positions of U2 samples. The other understanding tasks contribute through \cref{eq:sft-und}.

With the understanding path frozen, we initialize the generation tower from the base DiT and train it on G1 and G2, together with U3 annotation-generation data, using the supervised flow-matching loss of \cref{eq:sft-gen} (S1-3). We then continue SFT and match the frozen expert's velocity at states visited by the student's own denoising rollout~\cite{li2026diffusionopd} (S1-4),
\begin{equation}
  \label{eq:opd-gen}
  \mathcal{L}_{\mathrm{OPD}}^{\mathrm{gen}} = \mathbb{E}\,\bigl\| v_\theta\bigl(x_t^{\mathrm{roll}}, t\bigr) - v^{\mathrm{expert}}\bigl(x_t^{\mathrm{roll}}, t\bigr) \bigr\|_2^2.
\end{equation}
The expert is evaluated at the same state and timestep, with the expectation also taken over rollout noise, reflecting the importance of matched teacher-student information discussed in language-model policy distillation~\cite{wu2026dapd}. This loss is on-policy in state space, unlike the ground-truth-prefix understanding loss of \cref{eq:opd-und}.

Distillation supplements rather than replaces task-label supervision. S1-1 and S1-3 use SFT alone, while S1-2 and S1-4 combine SFT and distillation,
\begin{equation}
  \label{eq:combined}
  \mathcal{L}^{\mathrm{und}} = \alpha\,\mathcal{L}_{\mathrm{SFT}}^{\mathrm{und}} + \beta\,\mathcal{L}_{\mathrm{OPD}}^{\mathrm{und}},
  \;\;
  \mathcal{L}^{\mathrm{gen}} = \alpha\,\mathcal{L}_{\mathrm{SFT}}^{\mathrm{gen}} + \beta\,\mathcal{L}_{\mathrm{OPD}}^{\mathrm{gen}}.
\end{equation}
Here $\alpha,\beta\ge0$ are run-level weights. We use $(\alpha_1,\beta_1)$ for S1-2 and $(\alpha_2,\beta_2)$ for S1-4. Their values and ablations appear in \cref{sec:experiments,fig:ablation_supervision}.

\subsection{Parallel completion and assembly}
\label{sec:parallel_assembly}

After Stage~1, the understanding tower needs to reinforce U1 without losing its other tasks, while the generation tower needs further supervised and expert-guided updates. We therefore fork the S1-4 checkpoint and continue each tower in a separate branch with its own objective and schedule (\cref{fig:training_pipeline,alg:training_algorithm}). Branch~A freezes the generation tower and performs supervised understanding-task replay weighted toward U1 using \cref{eq:sft-und} (S2-1-1), while retaining the other understanding tasks in the replay mixture. Branch~B freezes the understanding tower and continues generation training with the combined objective of \cref{eq:combined}, using \cref{eq:sft-gen,eq:opd-gen} and the same $(\alpha_2,\beta_2)$ as S1-4 (S2-1-2). The separate branches avoid direct gradient competition on the tower parameters and allow different training lengths, while the generation branch still reads representations from the frozen understanding path.

We merge the understanding tower and shared modules from Branch~A with the generation tower and latent projection modules from Branch~B to obtain UniSTEM (S2-2). Both branches start from S1-4, and keeping the shared modules with Branch~A preserves their training state alongside the selected understanding tower. We evaluate the assembled checkpoint directly because Branch~B was trained with the S1-4 understanding path rather than Branch~A's final path. Assembly requires neither weight averaging nor another joint training run. \Cref{tab:training_stages} gives the full settings.

%% file: sec/5_experiments_arxiv.tex
\section{Experiments}
\label{sec:experiments}
\setcounter{topnumber}{4}
\setcounter{bottomnumber}{4}
\setcounter{totalnumber}{8}
\setcounter{dbltopnumber}{3}
\renewcommand{\topfraction}{0.95}
\renewcommand{\bottomfraction}{0.95}
\renewcommand{\dbltopfraction}{0.95}
\renewcommand{\textfraction}{0.05}

\subsection{Setup}
\label{sec:exp_setup}

We compare UniSTEM with five task experts and eight external models on the full structure-disjoint test sets of \cref{sec:dataset_scale}. Four experts serve individual tasks, while one serves both U4 and G1. External models are evaluated zero-shot without training on this benchmark. Training settings and external-model protocols are detailed in \cref{sec:supp_training_settings,sec:supp_competitors}.

\subsection{Main results}
\label{sec:main_results}

We evaluate detection with one-to-one matching and F1, structure prediction with formula accuracy and lattice/orientation errors, conductivity in the $\log_{10}$ domain, and generated images with SSIM~\cite{wang2004ssim} and PSNR.\footnote{Evaluation code: \url{https://github.com/leonardokong486/UniSTEM}. UniSTEM model weights: \url{https://huggingface.co/LeonardoKong/UniSTEM}. Complete test benchmark: \url{https://huggingface.co/datasets/LeonardoKong/UniSTEM-bench}.} A fixed pixel tolerance for vacancy detection varies with column spacing, while whole-map SSIM can be dominated by background. We therefore also report column-spacing-adaptive F1 for U1 and atomic-column-neighborhood SSIM for G1. Their calculation and visualizations are detailed in \cref{sec:supp_adaptive,sec:supp_adaptive_u1,sec:supp_adaptive_g1}. We additionally report direction-angle error for G1 and multiplicative conductivity error for U4. The U4 result in \cref{tab:main_results} uses ground-truth DPC input. Serial inference with reconstructed DPC is evaluated in \cref{tab:serial_inference_main}. \Cref{tab:main_results} reports representative full-test-set scores for all six tasks.

\begin{figure*}[!t]
  \centering
  \includegraphics[width=0.85\textwidth]{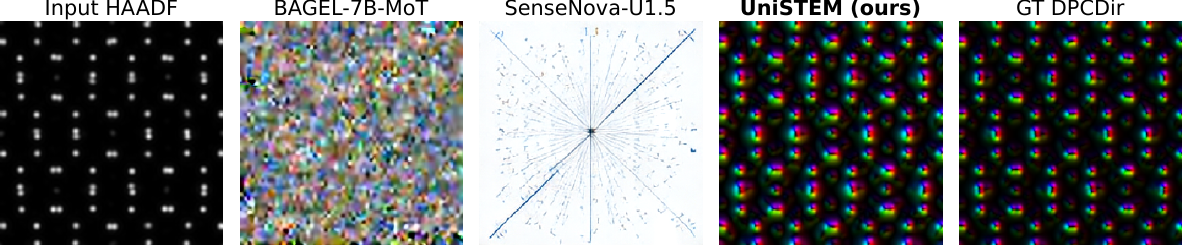}
  \caption{G1 DPC direction-field reconstruction from a HAADF micrograph. The panels compare UniSTEM and competing models with ground truth.}
  \label{fig:g1_visualization}
\end{figure*}

\begin{figure*}[!t]
  \centering
  \includegraphics[width=0.9\textwidth]{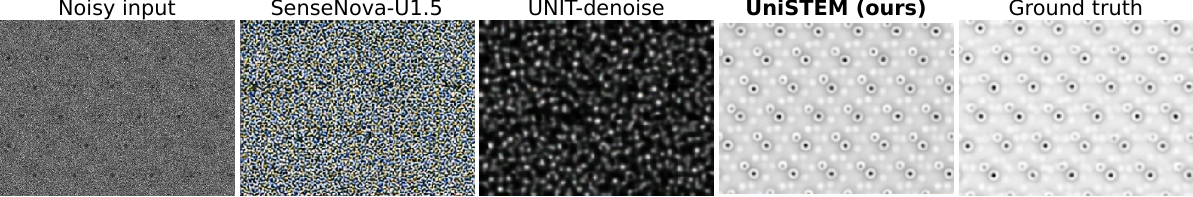}
  \caption{G2 denoising of an ABF micrograph at SNR 5. The panels compare the outputs of UniSTEM and competing models with ground truth.}
  \label{fig:g2_visualization}
\end{figure*}

\begin{table*}[t]
  \centering
  \caption{Full-test-set results. Expert denotes the corresponding task expert. External models are evaluated zero-shot on this benchmark. Arrows indicate the preferred direction. A dash denotes an unsupported task or unavailable metric. $\sigma_p$ uses ground-truth DPC input. Evaluation details and descriptions of the external models are given in \cref{sec:supp_metrics,sec:supp_competitors}.}
  \label{tab:main_results}
  \fontsize{7.5}{8.7}\selectfont
  \setlength{\tabcolsep}{2.5pt}
  \begin{tabular}{@{}lllcccccccccc@{}}
    \toprule
    \singlelinehead{\textbf{Task}} & \singlelinehead{\textbf{Metric}} & & \singlelinehead{\textbf{Expert}} & \singlelinehead{\textbf{UniSTEM}} & \textbf{\shortstack{BAGEL-\\7B}} & \singlelinehead{\textbf{ATOMIC}} & \textbf{\shortstack{Qwen3-\\VL-8B}} & \textbf{\shortstack{SenseNova-\\U1.5-8B}} & \textbf{\shortstack{InternVL\\3.5-8B}} & \textbf{\shortstack{Microscopy\\GPT}} & \textbf{\shortstack{UNIT-\\denoise}} & \singlelinehead{\textbf{tk\_r\_em}} \\
    \midrule
    \multirow{2}{*}{U1 Vacancy} & F1@5px & $\uparrow$ & 0.420 & 0.374 & 0.000 & 0.000 & 0.000 & 0.001 & 0.006 & - & - & - \\
                                & F1@adaptive ($k{=}0.5$) & $\uparrow$ & 0.591 & \textbf{0.643} & 0.002 & 0.000 & - & 0.011 & 0.012 & - & - & - \\
    \midrule
    \multirow{2}{*}{U2 Structure} & formula acc.\ & $\uparrow$ & 0.566 & 0.561 & 0.151 & 0.000 & 0.214 & 0.200 & 0.209 & 0.078 & - & - \\
                                   & lattice pooled rel.\ err.\ & $\downarrow$ & 0.0708 & \textbf{0.0582} & 0.1080 & - & 0.2135 & 0.2762 & 0.1159 & 0.0817 & - & - \\
    \midrule
    \multirow{2}{*}{U3 Dislocation} & F1@20px & $\uparrow$ & 0.675 & \textbf{0.720} & 0.000 & 0.000 & 0.000 & 0.000 & 0.000 & - & - & - \\
                                     & IoU F1@0.5 & $\uparrow$ & 0.522 & \textbf{0.560} & 0.000 & 0.000 & 0.000 & 0.000 & 0.000 & - & - & - \\
    \midrule
    \multirow{2}{*}{U4 Conductivity} & $\sigma_p$ log-MAE & $\downarrow$ & 0.772 & \textbf{0.650} & 2.751 & 9.7483 & 2.093 & 3.451 & 2.055 & - & - & - \\
                                      & $\sigma_p$ Spearman & $\uparrow$ & 0.799 & \textbf{0.848} & -0.049 & 0.1374 & 0.017 & - & 0.069 & - & - & - \\
    \midrule
    \multirow{3}{*}{G1 Field recon.} & DPCDir SSIM & $\uparrow$ & \textbf{0.896} & 0.862 & 0.026 & - & - & 0.058 & - & - & - & - \\
                                      & direction err.\ mean & $\downarrow$ & 17.44 & \textbf{15.20} & 90.90 & - & - & 109.20 & - & - & - & - \\
                                      & nb.\ SSIM (peak, r5) & $\uparrow$ & 0.845 & \textbf{0.855} & 0.042 & - & - & 0.027 & - & - & - & - \\
    \midrule
    \multirow{2}{*}{G2 Denoising} & PSNR (dB) & $\uparrow$ & 32.47 & \textbf{33.03} & 6.13 & - & - & 8.99 & - & - & 9.63 & 20.36 \\
                                   & SSIM & $\uparrow$ & \textbf{0.919} & 0.914 & 0.086 & - & - & 0.214 & - & - & 0.263 & 0.757 \\
    \bottomrule
  \end{tabular}
\end{table*}

\begin{figure*}[t]
  \centering
  \includegraphics[width=\textwidth]{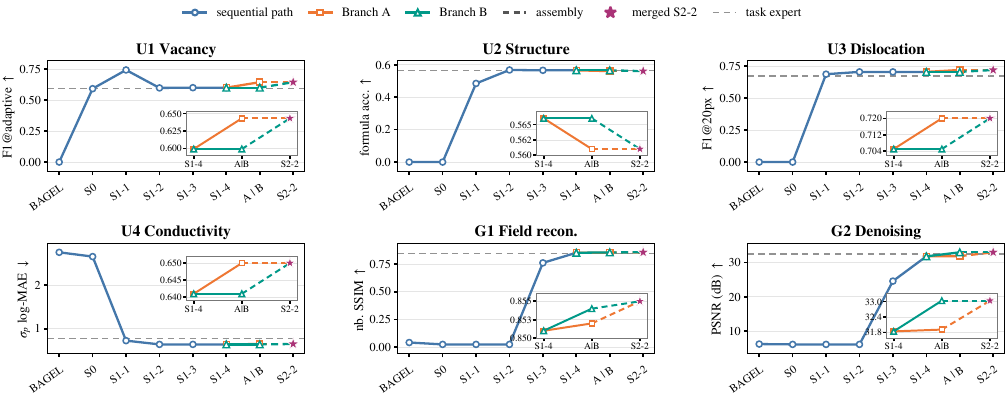}
  \caption{Six task metrics along the training path. U1 uses column adaptive F1. Dashed gray lines mark task experts. Insets show Stage~2 in the original metric units. Dashed lines leading to S2-2 indicate assembly. The generation tower has not been trained before S1-3.}
  \label{fig:stage_progression}
\end{figure*}

\begin{figure}[!htbp]
  \centering
  \includegraphics[width=0.98\columnwidth]{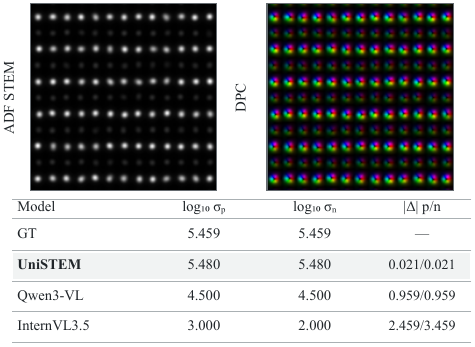}
  \caption{U4 conductivity prediction from an ADF STEM micrograph and its ground-truth DPC direction map. All models shown receive both images. Predictions and absolute errors are in the $\log_{10}$ domain.}
  \label{fig:u4_visualization}
\end{figure}

The external models struggle with atomic-scale grounding on this benchmark. Their vacancy and dislocation F1 scores are near zero, the best zero-shot formula accuracy is $0.214$, and the best $\sigma_p$ log-MAE is $2.055$. Models with image-generation capabilities fall short of UniSTEM on field reconstruction and denoising. Microscopy-specific models cover only their supported tasks, with MicroscopyGPT reaching $0.078$ formula accuracy and UNIT-denoise reaching $9.63$\,dB PSNR.

UniSTEM performs all six tasks with one model and surpasses the corresponding task experts on $9$ of the $13$ reported metrics. Relative to the experts, dislocation F1@20px improves by $6.7\%$, conductivity log-MAE falls by $15.8\%$, and DPC direction error falls by $12.8\%$. Some gaps remain, including an $11.0\%$ lower fixed-tolerance vacancy F1 and a $3.8\%$ lower whole-map DPC SSIM. These differences show that performance gains depend on the task and metric. Supplementary metrics beyond those selected for \cref{tab:main_results} are reported for U1--U4 and G1--G2 in \cref{sec:supp_u1,sec:supp_u2,sec:supp_u3,sec:supp_u4,sec:supp_g1,sec:supp_g2}. Error coupling and performance across imaging conditions are analyzed in \cref{sec:supp_coupling}.

\begin{table*}[!t]
  \centering
  \caption{Serial inference on $3{,}904$ shared test samples. U4 always receives G1-predicted DPC; denoised inputs first pass through G2. Parentheses give the retained percentage relative to the clean row: score/clean for SSIM and PSNR, and clean/error for log-MAE, so values above $100\%$ indicate improvement. PSNR percentages compare the reported dB values.}
  \label{tab:serial_inference_main}
  \footnotesize
  \setlength{\tabcolsep}{5pt}
  \begin{tabular}{@{}lcccc@{}}
    \toprule
    \singlelinehead{Input condition} & \shortstack{G1 DPC\\SSIM $\uparrow$} & \shortstack{G1 DPC\\PSNR (dB) $\uparrow$} & \shortstack{U4 $\sigma_p$\\log-MAE $\downarrow$} & \shortstack{U4 $\sigma_n$\\log-MAE $\downarrow$} \\
    \midrule
    Clean & 0.8616 (100.0\%) & 25.628 (100.0\%) & 0.6567 (100.0\%) & 0.6194 (100.0\%) \\
    $s=2$, noisy & 0.4634 (53.8\%) & 18.108 (70.7\%) & 0.7370 (89.1\%) & 0.6938 (89.3\%) \\
    $s=2$, denoised & \textbf{0.7651} (88.8\%) & \textbf{23.068} (90.0\%) & \textbf{0.6563} (100.1\%) & \textbf{0.6198} (99.9\%) \\
    $s=10$, noisy & 0.2689 (31.2\%) & 11.733 (45.8\%) & 0.6519 (100.7\%) & 0.6139 (100.9\%) \\
    $s=10$, denoised & \textbf{0.8282} (96.1\%) & \textbf{24.608} (96.0\%) & 0.6609 (99.4\%) & 0.6231 (99.4\%) \\
    \bottomrule
  \end{tabular}
\end{table*}

To test U4 with a generated rather than ground-truth DPC map, we feed it G1 outputs on the shared test set (\cref{tab:serial_inference_main}). On clean images, $\sigma_p$ log-MAE is $0.6567$, close to $0.650$ with ground-truth DPC in \cref{tab:main_results}. Under synthetic degradation, G2 improves G1 DPC SSIM at both settings. It lowers U4 log-MAE from $0.7370$ to $0.6563$ at $s=2$, but not at $s=10$ ($0.6519$ to $0.6609$). The latter case motivates reporting both field similarity and downstream error. The settings do not isolate noise variance because corruption also rescales intensity. All five conditions use the same 3{,}904 held-out G1/U4 samples. U4 receives G1-predicted DPC for every condition, while G2 precedes G1 and U4 in the denoised conditions. The corruption combines Poisson noise with per-image intensity rescaling. At $s=2$ and $s=10$, G2 raises G1 DPC reconstruction PSNR by $4.96$ and $12.88$\,dB, respectively. The differing U4 responses show that stronger field reconstruction does not always reduce downstream conductivity error.

\begin{figure}[!htbp]
  \centering
  \includegraphics[width=\columnwidth]{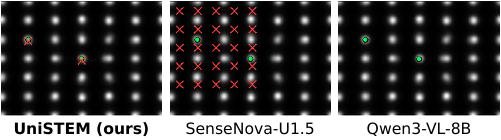}
  \caption{U1 vacancy localization. Green dots mark ground-truth vacancies and red crosses mark predicted sites.}
  \label{fig:u1_visualization}
\end{figure}

\subsection{Training progress and strategy ablations}
\label{sec:stage_dynamics}

\Cref{fig:stage_progression} tracks performance across the six tasks. Mixed understanding SFT raises U1 adaptive F1 by $25\%$ from S0 to S1-1. During S1-2, U2 formula accuracy increases by $17\%$, while U1 adaptive F1 falls by $19\%$. The gradient probe finds no significant directional conflict at the measured checkpoints (\cref{sec:supp_grad_probe}). Experiments varying data mixtures across training stages in \cref{sec:supp_ratio} suggest that sampling ratios partly explain the U1 performance decline. Subsequent generation training raises G1 neighborhood SSIM by $12\%$ and G2 PSNR by $7.25$\,dB from S1-3 to S1-4.

\begin{figure}[t]
  \centering
  \includegraphics[width=\columnwidth]{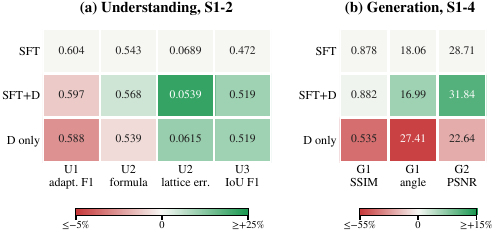}
  \caption{Supervision ablations on the full test sets. D denotes distillation. Cells show raw scores and color shows favorable relative change from SFT alone. The two panels use separate color scales. Lower lattice and direction errors are better.}
  \label{fig:ablation_supervision}
\end{figure}

\begin{figure}[!htbp]
  \centering
  \includegraphics[width=0.86\columnwidth]{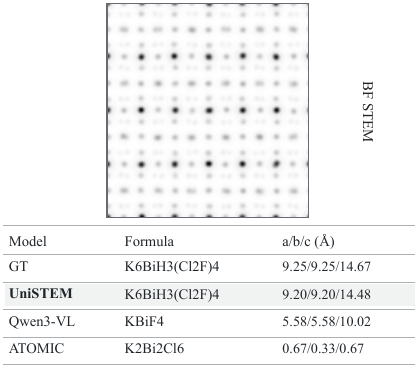}
  \caption{U2 structure prediction from a BF STEM micrograph. The figure shows the predicted formula and lattice lengths for one sample. ATOMIC values are transcribed from its raw response, which does not satisfy the benchmark output schema.}
  \label{fig:u2_visualization}
\end{figure}

\Cref{fig:ablation_supervision} tests the supervision choices in Stage~1. Targeted expert guidance raises U2 formula accuracy by $4.6\%$ and reduces its lattice error by $21.8\%$ relative to SFT alone. For generation, combining SFT with OPD raises G2 PSNR by $3.13$\,dB and reduces G1 direction error by $5.9\%$. Distillation alone substantially degrades generation quality. These comparisons support using expert guidance alongside task label supervision.

\begin{figure}[t]
  \centering
  \includegraphics[width=\columnwidth]{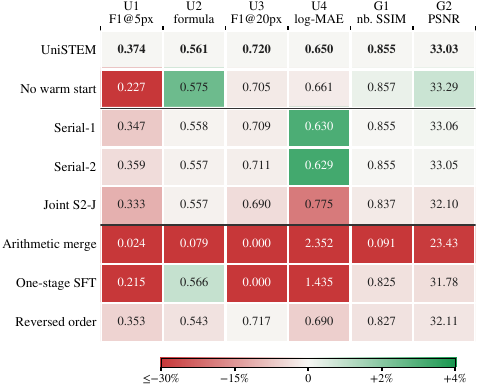}
  \caption{Representative full test set metrics for alternative training routes. Cells show raw scores and color shows favorable relative change from UniSTEM. Losses beyond $30\%$ use the darkest red. Lower U4 log MAE is better.}
  \label{fig:ablation_routes}
\end{figure}

\begin{figure}[!htbp]
  \centering
  \includegraphics[width=0.84\columnwidth]{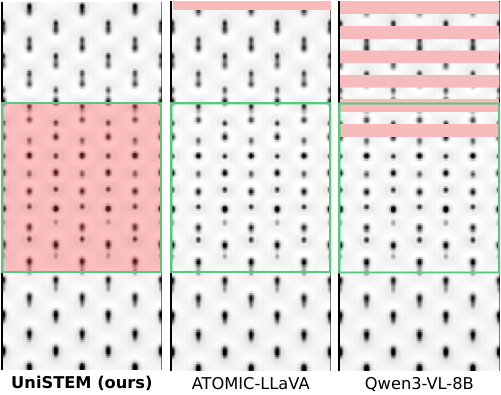}
  \caption{U3 dislocation localization. Green boxes mark the ground-truth region and red overlays mark model predictions.}
  \label{fig:u3_visualization}
\end{figure}

We also test the order of Stage~1 tower training in \cref{fig:ablation_routes}. Training generation before understanding still yields a functional model, but its final U1 F1 at $5$\,px is $5.6\%$ lower, U4 log MAE is $6.2\%$ higher, and U2 lattice error is $12.0\%$ higher than with the reported order. The reversed route remains close on most metrics, while these differences favor adapting understanding before generation.

In Stage~2, separate branches complete the two towers before assembly. The assembled model retains Branch~A's U1 performance and Branch~B's G2 performance. Relative to the better branch values, U2 formula accuracy falls by only $0.9\%$ and U4 log MAE rises by $1.4\%$. Joint training from the same S1-4 checkpoint with a comparable update budget performs worse than assembly on $11$ of $13$ metrics, including an $8.1\%$ reduction in U1 adaptive F1 and a $19.2\%$ increase in U4 log MAE. Completing the branches serially preserves generation quality but lowers U1 F1 at $5$\,px by $7.2\%$ or $4.0\%$, depending on the order. These results favor separate branch completion followed by assembly.

The U1 expert warm start also helps retain the selected localization capability. With all later training held fixed, removing this initialization lowers final U1 adaptive F1 by $24.9\%$ and F1 at $5$\,px by $39.3\%$. The warm start therefore supports the priority placed on U1, while its effects on the other tasks vary across metrics.

To study model performance under simpler multitask unification schemes, we compare task arithmetic merging and one stage mixed SFT with the full recipe in \cref{fig:ablation_routes}. Arithmetic merging yields zero U3 detection F1. Mixed SFT brings G2 PSNR within $1.25$\,dB of UniSTEM, but U3 detection F1 remains zero and U4 log MAE increases by $121\%$. Together with the Stage~1 order and Stage~2 completion ablations, these results support staged training, targeted expert guidance, and separate tower completion as components of the final training recipe.

\subsection{Visualization}
\label{sec:visualization}

\Cref{fig:g1_visualization,fig:g2_visualization} provide qualitative comparisons for the two generation tasks. In the G1 DPC field example, BAGEL-7B-MoT produces noise without the periodic field structure, while SenseNova-U1.5 hallucinates a pronounced radial texture absent from the ground truth. UniSTEM instead recovers the atomic-column field pattern close to the target. In the G2 denoising example, UNIT-denoise turns many background noise fluctuations into atom-like spots, and SenseNova-U1.5 leaves most of the noise intact. UniSTEM preserves the ring-like atomic-column contrast of the clean target without introducing such spurious spots. These observations concern the displayed samples. The full-test-set comparison is reported in \cref{tab:main_results}. Additional denoising comparisons on real, experimentally acquired HAADF images are given in \cref{sec:supp_real_denoising}.

\Cref{fig:u3_visualization} compares dislocation localization. UniSTEM covers the annotated region, while both general-purpose models predict regions elsewhere.
\Cref{fig:u1_visualization} shows vacancy localization on a common input. UniSTEM places its predictions at the annotated sites, whereas SenseNova-U1.5 produces many false positives. Qwen3-VL-8B, ATOMIC-LLaVA, and InternVL3.5-8B produce no valid coordinates for this example.

For U2, \cref{fig:u2_visualization} displays selected structured outputs beside the input micrograph. The element set is supplied to the models as part of the task prompt. UniSTEM recovers the exact formula $\mathrm{K_6BiH_3(Cl_2F)_4}$ and estimates the three lattice lengths within $0.19$\,\AA{} of the ground truth in this example. Qwen3-VL-8B predicts a different stoichiometry and shorter lattice lengths. ATOMIC's raw response also gives a different formula and cannot be scored under the required JSON schema. This example illustrates the formula and lattice differences summarized over the full test set in \cref{tab:main_results}.

For U4, \cref{fig:u4_visualization} uses the same oracle-input setting as the main results, with the original STEM micrograph and ground-truth DPC map supplied together. UniSTEM predicts both $\log_{10}\sigma_p$ and $\log_{10}\sigma_n$ within $0.021$ of the ground truth for this sample. Qwen3-VL-8B gives an error of $0.959$ for each value, while InternVL3.5-8B gives errors of $2.459$ and $3.459$. The example shows the quantitative difference between the models on one material and complements the full-test-set U4 results in \cref{tab:main_results}.

%% file: sec/conclusion.tex
\section{Conclusion}

We introduce a six-task STEM benchmark spanning atomic-resolution understanding, property prediction, and image generation. UniSTEM combines targeted expert guidance with separate tower optimization and assembly, enabling structured predictions and spatial image generation in one model. On structure-disjoint test sets, it exceeds task experts on nine of thirteen reported metrics, although fixed-pixel vacancy F1 and whole-map DPC SSIM remain below their experts. Serial inference shows that improved field reconstruction need not reduce conductivity error. Registered experimental measurements could directly test error propagation and scientific reliability.

%% file: sec/6_appendix_arxiv.tex
\appendix
\section{Benchmark and Training Details}
\label{sec:supp_setup}

\subsection{Benchmark Size}
\label{sec:supp_benchmark_size}

\Cref{tab:scale} gives the processed sample counts, and \cref{tab:training_stages} records the data mixtures, losses, and schedules for the training runs.

\begin{table}[!htbp]
  \centering
  \caption{Processed training and test samples. G1 and U4 share one corpus.}
  \label{tab:scale}
  \small
  \setlength{\tabcolsep}{5pt}
  \begin{tabular}{@{}lrrr@{}}
    \toprule
    \textbf{Tasks} & \textbf{Train} & \textbf{Test} & \textbf{Test mats.} \\
    \midrule
    G1 + U4 (field + conductivity) & 73{,}956 & 3{,}904 & 976 \\
    G2 (denoising) & 303{,}552 & 16{,}048 & 1{,}003 \\
    U1 (vacancy) & 45{,}592 & 2{,}484 & 56 \\
    U2 (structure) & 83{,}347 & 4{,}392 & 777 \\
    U3 (dislocation) & 5{,}371 & 300 & 76 \\
    \bottomrule
  \end{tabular}
\end{table}

\Cref{tab:test_condition_counts} summarizes the imaging conditions and other slice axes of the test sets. The four detector modes have nearly equal counts in each corpus. G1 and U4 use the same $3{,}904$ test images. We divide recorded defocus values at $-10$\,\AA{} into negative defocus (neg, $<-10$\,\AA) and near-zero defocus (near0, $\geq-10$\,\AA). The U2 zone-axis counts use dataset directory labels, which differ from the ground-truth \texttt{z\_orientation} vector categories discussed in the orientation analysis below.

\subsection{Trainable and Frozen Components in Dual-Tower Training}
\label{sec:supp_tower_paths}

\begin{figure}[!htbp]
  \centering
  \includegraphics[width=\textwidth]{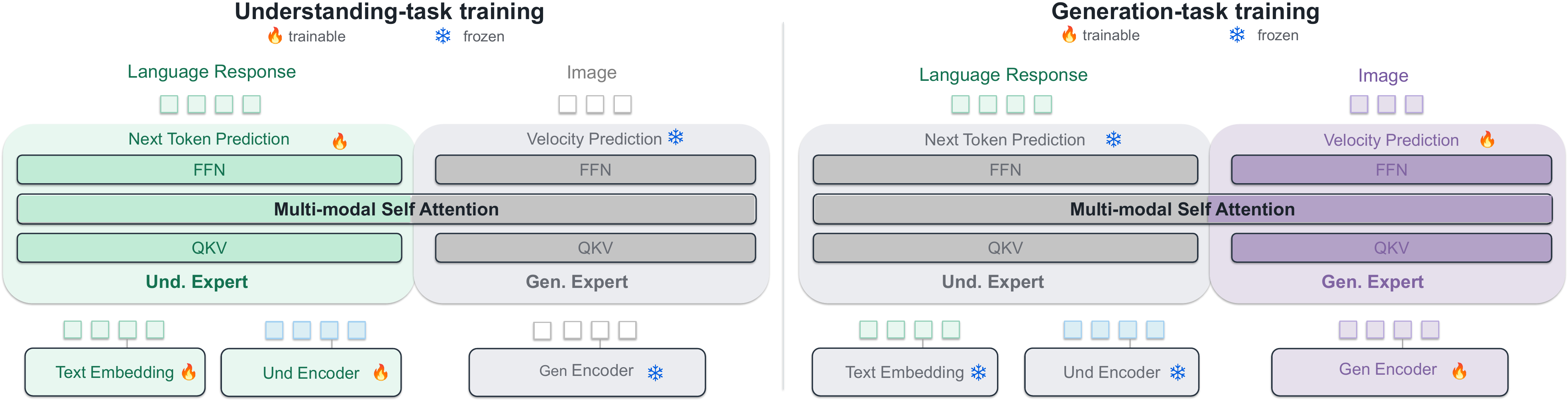}
  \caption{Trainable and frozen components in the two tower-specific training configurations. The left panel shows understanding-tower training and the right panel shows generation-tower training. Snowflakes and flames mark frozen and trainable modules, respectively.}
  \label{fig:dual_tower_sep_train}
\end{figure}

\Cref{fig:dual_tower_sep_train} shows the two training configurations, and \cref{tab:training_stages} identifies the stages using each one. The 28-layer Transformer uses joint attention and routes tokens through the understanding or generation tower according to token type. The understanding-tower configuration is used in S1-1, S1-2, and Stage~2 Branch~A (S2-1-1). The generation-tower configuration is used in S1-3, S1-4, and Stage~2 Branch~B (S2-1-2). Parameter counts below are measured from training logs.

\noindent\textbf{Understanding-tower configuration.} The ViT and connector encode the input micrograph, while \texttt{embed\_tokens} encodes the text. The resulting image and text tokens form a packed sequence without latent tokens. They enter the understanding tower, and \texttt{lm\_head} produces the output text tokens. The ViT, connector, token embeddings, understanding tower, and language head are trainable, totaling $8{,}035.2$M parameters. The generation tower receives no tokens and does not participate in the forward pass. It is not loaded in S1-1 or S1-2, whereas Branch~A retains it in the model with its parameters frozen during replay. The sinusoidal \texttt{vit\_pos\_embed} remains frozen, and the VAE does not participate in this configuration. With the frozen generation tower retained, the training logs report $6{,}572.1$M frozen parameters.

\noindent\textbf{Generation-tower configuration.} The frozen VAE encoder maps images to latents, and frozen token embeddings encode the text prompt. G1 additionally uses the frozen ViT and connector to encode an image condition, whereas G2 does not use this image-conditioning branch. Text tokens, available image-condition tokens, and noisy latent tokens enter the same Transformer. The frozen understanding tower encodes the text and image-condition tokens, providing keys and values that the generation tower can read. Noisy latent tokens enter the generation tower through \texttt{vae2llm}, latent positional embeddings, and timestep embeddings. The generation output is mapped to the latent-space prediction through \texttt{llm2vae}. The generation tower and these latent-interface modules are trainable, totaling $6{,}554.5$M parameters. The ViT, connector, token embeddings, understanding tower, and \texttt{lm\_head} remain frozen, totaling $8{,}052.7$M frozen parameters. Training does not require VAE decoding.

\subsection{Training Settings}
\label{sec:supp_training_settings}

\begin{table}[!htbp]
  \centering
  \caption{Training stages and settings. Und/Gen indicate the trained tower, with the other tower frozen or omitted as applicable. SFT uses task labels and OPD uses expert guidance.}
  \label{tab:training_stages}
  \small
  \setlength{\tabcolsep}{3.5pt}
  \begin{tabular}{@{}llllll@{}}
    \toprule
    \textbf{Run} & \textbf{Tower} & \textbf{Init} & \textbf{Data (weights)} & \textbf{Supervision} & \textbf{Steps / lr} \\
    \midrule
    \multicolumn{6}{l}{\emph{Preparation --- expert training and warm start}} \\
    Experts & per-task & base & per-task corpus & $\mathcal{L}_{\mathrm{SFT}}^{\mathrm{und}/\mathrm{gen}}$ & per-task (supp.) \\
    S0 & warm start & U1 expert & --- & --- & --- \\
    \midrule
    \multicolumn{6}{l}{\emph{Stage 1 --- sequential single-tower training}} \\
    S1-1 & Und & S0 & U1 2.0, others 1.0 & $\mathcal{L}_{\mathrm{SFT}}^{\mathrm{und}}$ & 10k / 2e-5 \\
    S1-2 & Und & S1-1 & U1 1.0, others 1.0 & $\alpha_1\mathcal{L}_{\mathrm{SFT}}^{\mathrm{und}}{+}\beta_1\mathcal{L}_{\mathrm{OPD}}^{\mathrm{und}}$ & 5k / 1e-5 \\
    S1-3 & Gen & Und$=$S1-2, Gen$=$base & G1 + G2 60k + U3 & $\mathcal{L}_{\mathrm{SFT}}^{\mathrm{gen}}$ & 10k / 1e-5 \\
    S1-4 & Gen & S1-3 & G1 + G2 60k & $\alpha_2\mathcal{L}_{\mathrm{SFT}}^{\mathrm{gen}}{+}\beta_2\mathcal{L}_{\mathrm{OPD}}^{\mathrm{gen}}$ & 2k / 1e-5 \\
    \midrule
    \multicolumn{6}{l}{\emph{Stage 2 --- parallel completion and tower merging}} \\
    S2-1-1 (Br.~A) & Und & S1-4 & U1 2.0, others 0.4/0.2/0.4 & $\mathcal{L}_{\mathrm{SFT}}^{\mathrm{und}}$ & 2k / 1e-5 \\
    S2-1-2 (Br.~B) & Gen & S1-4 & G1 + G2 60k & $\alpha_2\mathcal{L}_{\mathrm{SFT}}^{\mathrm{gen}}{+}\beta_2\mathcal{L}_{\mathrm{OPD}}^{\mathrm{gen}}$ & 3k / 1e-5 \\
    S2-2 & merge & Und$\leftarrow$S2-1-1, Gen$\leftarrow$S2-1-2 & --- & --- & --- \\
    \bottomrule
  \end{tabular}
\end{table}

All six UniSTEM training runs use 8$\times$H200 GPUs with PyTorch fully-sharded data parallelism in the full-shard configuration~\cite{zhao2023pytorch}. Gradient accumulation is one, with one pack per GPU at each step. The packer uses a soft target of $4{,}096$ tokens per GPU for S1-1, S1-2, and S2-1-1, and $7{,}168$ for S1-3, S1-4, and S2-1-2. A pack has a hard limit of $8{,}192$ tokens. The number of samples per step therefore varies with their sequence lengths.

All runs optimize only the trainable parameters with AdamW~\cite{loshchilov2017fixing}, using momentum coefficients $(0.9,0.95)$, $\epsilon{=}10^{-15}$, zero weight decay, gradient-norm clipping at $1.0$, and seed $4396$. The learning rates in \cref{tab:training_stages} are constant after linear warmup. S1-1 warms up for $200$ steps to $2\mathrm{e}{-5}$, while the other five runs warm up for $100$ steps to $1\mathrm{e}{-5}$. No learning-rate decay follows warmup.

The task weights in \cref{tab:training_stages} define normalized probabilities for selecting task groups rather than proportions used to construct a fixed mixed dataset. S1-1 and Branch~A select groups at the sample level and allow mixed-task packs. S1-2, S1-4, and Branch~B select one group per pack, keeping each pack in a single domain for teacher guidance. S1-3 is the exception, reading its three tasks from one fixed shuffled stream. It uses all $73{,}956$ G1 training records, a fixed subset of approximately $60{,}000$ G2 records, and all $5{,}371$ U3 annotation-generation records. The same G2 subset is reused in S1-4 and Branch~B without resampling its membership between epochs.

Checkpoint handoffs use the U1 expert's EMA weights at step $6{,}500$ to initialize S1-1, S1-1 EMA weights to initialize S1-2, S1-2 raw weights to initialize S1-3, S1-3 EMA weights to initialize S1-4, and S1-4 raw weights to initialize both Stage~2 branches. EMA uses decay $0.9999$ in S1-1 and S1-3. The final model combines the raw understanding tower and shared modules from Branch~A at step $2{,}000$ with the raw generation tower and latent-interface modules from Branch~B at step $3{,}000$. Assembly copies these modules without weight averaging. The five preparation experts use pure SFT on their task data and are frozen after training. Tiny-set comparisons inform the U1-expert warm start and identify tasks for targeted distillation.

For the combined losses in \cref{eq:combined}, S1-2 uses $(\alpha_1,\beta_1){=}(1,1)$, applying ground-truth cross-entropy to all four understanding tasks and reverse-KL distillation only to U2 batches. S1-4 and S2-1-2 (Branch~B) use $(\alpha_2,\beta_2){=}(0.2,0.8)$, combining ground-truth flow-matching MSE with expert velocity matching. S1-1, S1-3, and S2-1-1 (Branch~A) use SFT alone. The joint-training ablation S2-J of \cref{fig:ablation_routes} retains $(\alpha_2,\beta_2){=}(0.2,0.8)$ on generation batches and uses pure ground-truth cross-entropy on understanding replay batches.

\begin{table}[!htbp]
  \centering
  \caption{Test-set sample counts by imaging condition and other slice axes. Slash-separated counts follow the category order in each column heading. Dashes mark axes unavailable for a task.}
  \label{tab:test_condition_counts}
  \scriptsize
  \setlength{\tabcolsep}{3pt}
  \resizebox{\textwidth}{!}{%
  \begin{tabular}{@{}lrlcllll@{}}
    \toprule
    \textbf{Task} & \textbf{Test} & \textbf{BF/ABF/ADF/HAADF} & \textbf{200/300\,kV} & \textbf{neg/near0} & \textbf{Zone 001/110/111/210/211} & \textbf{Elements 1/2/$\geq$3} & \textbf{SNR 2/5/10/20} \\
    \midrule
    U1 & 2{,}484 & 621/621/621/621 & 1{,}144/1{,}340 & 1{,}544/940 & - & - & - \\
    U2 & 4{,}392 & 1{,}098/1{,}098/1{,}098/1{,}098 & 2{,}236/2{,}156 & 1{,}552/2{,}840 & 3{,}108/412/172/224/476 & 100/1{,}376/2{,}916 & - \\
    U3 & 300 & 76/72/76/76 & 132/160 & 136/156 & - & - & - \\
    U4 & 3{,}904 & 976/976/976/976 & 2{,}168/1{,}736 & 1{,}980/1{,}924 & - & 52/1{,}112/2{,}740 & - \\
    G1 & 3{,}904 & 976/976/976/976 & 2{,}168/1{,}736 & 1{,}980/1{,}924 & - & 52/1{,}112/2{,}740 & - \\
    G2 & 16{,}048 & 4{,}012/4{,}012/4{,}012/4{,}012 & - & - & - & - & 4{,}012/4{,}012/4{,}012/4{,}012 \\
    \bottomrule
  \end{tabular}}
\end{table}

\subsection{External Models and Evaluation Protocols}
\label{sec:supp_competitors}

We evaluate eight external models without adapting them to the benchmark training data. Unsupported tasks or unavailable metrics are marked with a dash in \cref{tab:main_results}. For U2, BAGEL-7B, Qwen3-VL-8B, SenseNova-U1.5-8B-MoT, and InternVL3.5-8B-Instruct use clarified output-format prompts requiring the zone-axis orientation as an integer list. BAGEL-7B additionally requires a single compact JSON object with an explicit field schema because its original answers do not close within the generation budget. These clarifications constrain answer format without providing material composition, prediction hints, or worked examples. The prompts for the other five tasks are unchanged.

BAGEL-7B is the pretrained BAGEL-7B-MoT backbone used by UniSTEM and is evaluated zero-shot on all six tasks~\cite{deng2025bagel}. SenseNova-U1.5-8B-MoT is likewise evaluated on all six tasks~\cite{sensenova2026sensenovau1.5}. For U1--U4, SenseNova uses direct answering with its default chain-of-thought disabled. For G1 and G2, it uses its native image-editing path with the official defaults of 50 steps and classifier-free guidance scale 4.0. Qwen3-VL-8B~\cite{bai2025qwen3} and InternVL3.5-8B-Instruct~\cite{wang2025internvl3_5} are evaluated on U1--U4 only because they do not provide the image-decoding output required for G1 or G2. InternVL3.5-8B uses direct answering. ATOMIC~\cite{tu2026atomic} is evaluated on the understanding tasks for which its interface produces scorable outputs. MicroscopyGPT~\cite{choudhary2025microscopygpt}, a Llama-3.2-11B-Vision model with a LoRA adapter trained on C2DB formula-to-structure captions, is evaluated on U2 alone. For MicroscopyGPT, the parser also accepts a quoted integer orientation list such as \texttt{``[0, 0, 1]''}, while the remaining scoring rules are unchanged.

The publicly released UNIT-for-STEM-denoise checkpoint~\cite{wu2026unsupervised}, abbreviated as ``UNIT-denoise,'' and the published \emph{tk\_r\_em} single-shot electron-microscopy restoration network~\cite{lobato2024deep} are evaluated zero-shot on G2 only. For \texttt{tk\_r\_em}, we use the high-resolution STEM checkpoint (\texttt{sfr\_hrstem}) at its native $256\times256$ input size, with stride-$128$ overlapping tiles and blended outputs. The restored images are scored as produced, without retraining or additional intensity rescaling.

\section{In-depth Evaluation and Metric Analysis}

\subsection{Supplementary Metric Tables}
\label{sec:supp_metrics}

The main results table summarizes each task with a small set of representative metrics, but these scores cannot fully show how localization depends on the matching tolerance, how accurately each component of a structure is predicted, or how errors vary across evaluation conditions. To complement \cref{tab:main_results}, this subsection reports full-test-set metrics for tolerance sweeps, output parsing and structure prediction, quantitative error distributions, local reconstruction quality, and denoising at different noise levels. These results provide a more detailed view of the capabilities and failure modes of UniSTEM, the task experts, and applicable external models. Model sources, supported tasks, and inference settings are described in \cref{sec:supp_competitors}.

\subsubsection{U1 Vacancy Detection}
\label{sec:supp_u1}

\Cref{tab:supp_u1} reports tighter fixed-tolerance scores, a sweep of column-spacing-adaptive tolerances, and the mean absolute vacancy-count error. Fixed pixel radii probe coordinate precision, while the adaptive radii scale with the estimated column spacing of each image. The spacing estimator, matching rule, and geometric rationale are detailed in \cref{sec:supp_adaptive_u1}.

\begin{table}[!htbp]
  \centering
  \caption{Supplementary U1 vacancy-detection metrics on the full test set (2{,}484 samples). Scores already reported in \cref{tab:main_results} are omitted. Adaptive F1 uses $R{=}k\,d$ with one estimated spacing $d$ per image. The Qwen3-VL-8B and InternVL3.5-8B sweeps use their saved coordinate lists and include 2{,}480 images each, since four images lack a stable spacing peak.}
  \label{tab:supp_u1}
  \small
  \setlength{\tabcolsep}{3.5pt}
  \begin{tabular}{@{}lccccccc@{}}
    \toprule
    \textbf{Metric} & \textbf{Expert} & \textbf{UniSTEM} & \textbf{BAGEL-7B} & \textbf{ATOMIC} & \textbf{\shortstack{Qwen3-VL-\\8B}} & \textbf{\shortstack{SenseNova-\\U1.5-8B}} & \textbf{\shortstack{InternVL\\3.5-8B}} \\
    \midrule
    \textit{Fixed tolerance} & & & & & & & \\
    F1@1px $\uparrow$ & 0.147 & 0.095 & 0.000 & 0.000 & 0.000 & 0.000 & 0.001 \\
    F1@2px $\uparrow$ & 0.246 & 0.184 & 0.000 & 0.000 & 0.000 & 0.000 & 0.003 \\
    F1@3px $\uparrow$ & 0.313 & 0.263 & 0.000 & 0.000 & 0.000 & 0.001 & 0.004 \\
    \midrule
    \textit{Column-adaptive tolerance} & & & & & & & \\
    Adaptive F1 ($k{=}0.25$) $\uparrow$ & 0.479 & 0.477 & 0.000 & 0.000 & 0.000 & 0.003 & 0.007 \\
    Adaptive F1 ($k{=}0.35$) $\uparrow$ & 0.537 & 0.576 & 0.001 & 0.000 & 0.000 & 0.006 & 0.009 \\
    \midrule
    Count error (mean abs.) $\downarrow$ & 0.164 & 0.128 & 1.994 & 1.883 & 2.207 & 5.684 & 2.932 \\
    \bottomrule
  \end{tabular}
\end{table}

The supplementary sweep separates stringent coordinate localization from column-level detection. UniSTEM trails the task expert at 1--3\,px, is nearly tied at $k{=}0.25$ (0.477 versus 0.479), and exceeds the expert by $7.3\%$ at $k{=}0.35$ (0.576 versus 0.537). Its count error is $22.0\%$ lower (0.128 versus 0.164). The applicable external models remain near zero across the adaptive sweep, showing that producing vacancy coordinates does not by itself yield reliable localization.

\subsubsection{U2 Structure Prediction}
\label{sec:supp_u2}

The U2 prompt supplies constituent elements and a simulation zone-axis directory label (e.g., \texttt{001}), but not stoichiometric ratios, lattice parameters, or the target \texttt{z\_orientation} vector. The label need not match that vector. G1/U4 prompts also supply constituent elements. \Cref{tab:supp_u2} distinguishes compliance with supplied context from predicted structure; orientation errors should be read with this zone-axis context. Scores from \cref{tab:main_results} are omitted.

\begin{table}[!htbp]
  \centering
  \caption{Supplementary U2 metrics on the full test set (4{,}392 samples).}
  \label{tab:supp_u2}
  \small
  \setlength{\tabcolsep}{3pt}
  \begin{tabular}{@{}lcccccccc@{}}
    \toprule
    \textbf{Metric} & \textbf{Expert} & \textbf{UniSTEM} & \textbf{BAGEL-7B} & \textbf{ATOMIC} & \textbf{\shortstack{Qwen3-VL-\\8B}} & \textbf{\shortstack{SenseNova-\\U1.5-8B}} & \textbf{\shortstack{InternVL\\3.5-8B}} & \textbf{\shortstack{Microscopy\\GPT}} \\
    \midrule
    JSON parse rate $\uparrow$ & 1.0000 & 0.9993 & 1.0000 & 0.0000 & 1.0000 & 0.9964 & 0.9483 & 0.9984 \\
    Element-set acc.\ $\uparrow$ & 0.9989 & 0.9973 & 0.9481 & 0.0000 & 0.8427 & 0.7678 & 0.8867 & 0.5995 \\
    Element ratio TV $\downarrow$ & 0.0891 & 0.0902 & 0.2092 & - & 0.1838 & 0.2120 & 0.1789 & 0.3069 \\
    Lattice per-sample med.\ $\downarrow$ & 0.0067 & \textbf{0.0051} & 0.2740 & - & 0.1482 & 0.2084 & 0.1980 & 0.2189 \\
    Orientation err.\ mean ($^{\circ}$) $\downarrow$ & 4.659 & \textbf{3.854} & 11.183 & - & 6.888 & 29.710 & 26.499 & 0.000 \\
    \bottomrule
  \end{tabular}
\end{table}

For MicroscopyGPT, the parser also accepts the orientation list as a quoted string without changing any other scoring rule. The orientation errors of MicroscopyGPT and BAGEL-7B are strongly influenced by their frequent $[0,0,1]$ predictions and should not be read as evidence of image-grounded orientation recognition.

Element-set accuracy is primarily a measure of whether a model follows the instruction to retain the element identities already given in the prompt. It is not evidence that the model inferred those elements from the micrograph. UniSTEM reaches 0.9973, close to the task expert's 0.9989, indicating that specialization for electron microscopy has preserved this general instruction-following ability. General-purpose BAGEL-7B and Qwen3-VL-8B score higher on this measure than microscopy-specific MicroscopyGPT. ATOMIC has a zero JSON parse rate, so its zero element-set score cannot by itself distinguish failure to follow the element instruction from failure to produce a scorable answer. The more material-specific quantities remain informative. UniSTEM's element-ratio total variation is 0.0902, close to the expert's 0.0891, and its per-sample median lattice error is 0.0051 versus the expert's 0.0067. The external models generally show larger errors on these quantities even when they preserve the supplied element identities.

This parser recovers 182 additional parseable MicroscopyGPT samples without regenerating answers, but only five of those contain an exact formula. The extra valid parses are therefore mostly incorrect structural predictions rather than recovered task successes.

Orientation error requires particular care. MicroscopyGPT outputs $[0,0,1]$ in 4{,}090 of its 4{,}385 parseable answers, while that orientation is the most common ground truth, appearing in 3{,}228 of 4{,}392 samples. BAGEL-7B outputs the same vector in 3{,}891 of 4{,}392 answers. Four-index ground-truth orientations are excluded from this angular metric. A related prior appears in MicroscopyGPT's lattice estimates. Its median errors for $\alpha$ and $\beta$ are below $10^{-4}$, whereas the median relative errors for $a$, $b$, and $c$ are approximately $0.45$, $0.45$, and $0.68$. Reproducing common angles does not imply accurate recovery of lattice lengths.

Formula prediction also deteriorates as composition becomes more complex. Among one-element formulas, MicroscopyGPT predicts 29 of 100 exactly and BAGEL-7B predicts 88 of 100. For formulas with four or more elements, those counts fall to 0 of 663 and 1 of 664, respectively, with the two- and three-element groups lying between these extremes. This stratification shows that a valid output format and correct use of the supplied element list do not establish an ability to infer multi-element stoichiometry from the image.

\begin{table}[!htbp]
  \centering
  \caption{Supplementary U3 dislocation-detection metrics on the full test set (300 samples). Scores already reported in \cref{tab:main_results} are omitted.}
  \label{tab:supp_u3}
  \small
  \setlength{\tabcolsep}{3.5pt}
  \begin{tabular}{@{}lccccccc@{}}
    \toprule
    \textbf{Metric} & \textbf{Expert} & \textbf{UniSTEM} & \textbf{BAGEL-7B} & \textbf{ATOMIC} & \textbf{\shortstack{Qwen3-VL-\\8B}} & \textbf{\shortstack{SenseNova-\\U1.5-8B}} & \textbf{\shortstack{InternVL\\3.5-8B}} \\
    \midrule
    F1@5px $\uparrow$ & 0.3313 & \textbf{0.3685} & 0.0000 & 0.0000 & 0.0000 & 0.0000 & 0.0000 \\
    F1@10px $\uparrow$ & 0.5103 & 0.5309 & 0.0000 & 0.0000 & 0.0000 & 0.0000 & 0.0000 \\
    F1@30px $\uparrow$ & 0.7570 & \textbf{0.7612} & 0.0000 & 0.0000 & 0.0041 & 0.0000 & 0.0000 \\
    Count error (mean abs.) $\downarrow$ & 0.0167 & 0.0100 & 1.3833 & 1.7500 & 1.4667 & 1.3733 & 1.6167 \\
    \bottomrule
  \end{tabular}
\end{table}

\subsubsection{U3 Dislocation Detection}
\label{sec:supp_u3}

\Cref{tab:supp_u3} supplements the main results with stricter and wider pixel tolerances and the mean absolute dislocation-count error. These measures separate coordinate precision from counting accuracy across the 300 test images.

UniSTEM exceeds the task expert at the additional 5, 10, and 30\,px tolerances, with F1 scores of 0.3685 versus 0.3313, 0.5309 versus 0.5103, and 0.7612 versus 0.7570, respectively. These correspond to relative gains of $11.2\%$, $4.0\%$, and $0.6\%$, so the advantage is most visible under stricter localization. UniSTEM also has a lower count error, 0.0100 versus 0.0167. The external models remain near zero even at the wider tolerance, indicating that their outputs do not reliably localize dislocations.

\subsubsection{U4 Conductivity Prediction}
\label{sec:supp_u4}

All models use the ground-truth DPC map as input. The supplementary metrics in \cref{tab:supp_u4} examine large errors, variation across samples, and multiplicative deviations from the true conductivity.

\begin{table}[!htbp]
  \centering
  \caption{Supplementary U4 conductivity metrics on the full test set (3{,}904 samples, $\log_{10}$ domain, ground-truth DPC input). Scores already reported in \cref{tab:main_results} are omitted. $F=10^{|\log_{10}\hat\sigma-\log_{10}\sigma|}$ is the multiplicative error factor, and ``$F{\le}c\times$'' is the fraction of predictions within a factor $c$ of the truth. SenseNova-U1.5-8B's $\sigma_p$ Pearson coefficient is undefined because its predictions are nearly constant. ATOMIC yields a parsable conductivity value on 3{,}514 of the 3{,}904 test images, and its entries are computed on those images.}
  \label{tab:supp_u4}
  \small
  \setlength{\tabcolsep}{3.5pt}
  \begin{tabular}{@{}lccccccc@{}}
    \toprule
    \textbf{Metric} & \textbf{Expert} & \textbf{UniSTEM} & \textbf{BAGEL-7B} & \textbf{ATOMIC} & \textbf{\shortstack{Qwen3-VL-\\8B}} & \textbf{\shortstack{SenseNova-\\U1.5-8B}} & \textbf{\shortstack{InternVL\\3.5-8B}} \\
    \midrule
    $\sigma_p$ log-RMSE $\downarrow$ & 1.3676 & \textbf{1.2285} & 3.4225 & 11.2521 & 2.5862 & 4.0599 & 2.4737 \\
    $\sigma_p$ $R^2$ $\uparrow$ & 0.5913 & \textbf{0.6636} & -1.6512 & -26.7281 & -0.4616 & -2.6018 & -0.3433 \\
    $\sigma_p$ Pearson $\uparrow$ & 0.8064 & \textbf{0.8356} & -0.0493 & 0.1616 & -0.0232 & - & 0.0703 \\
    $\sigma_n$ log-MAE $\downarrow$ & 0.7046 & \textbf{0.6146} & 2.9748 & 9.9848 & 1.7314 & 2.7673 & 1.8345 \\
    $\sigma_n$ $R^2$ $\uparrow$ & 0.5278 & \textbf{0.6082} & -2.4849 & -36.1853 & -0.0653 & -2.1720 & -0.4079 \\
    $\Delta{=}\log\sigma_p{-}\log\sigma_n$ MAE $\downarrow$ & 0.2481 & \textbf{0.2153} & 0.4893 & 0.3210 & 0.9215 & 0.7080 & 1.3728 \\
    \midrule
    \textit{Multiplicative error factor $F$ for $\sigma_p$} & & & & & & & \\
    $F$ median $\downarrow$ & 1.996 & \textbf{1.737} & 806.439 & $9.7{\times}10^{11}$ & 36.550 & 881.760 & 82.050 \\
    $F{\le}1.25\times$ $\uparrow$ & 0.2205 & \textbf{0.2379} & 0.0335 & 0.0017 & 0.0277 & 0.0020 & 0.0192 \\
    $F{\le}1.5\times$ $\uparrow$ & 0.3671 & \textbf{0.4055} & 0.0508 & 0.0020 & 0.0553 & 0.0020 & 0.0427 \\
    $F{\le}2\times$ $\uparrow$ & 0.5015 & \textbf{0.5727} & 0.0822 & 0.0031 & 0.1004 & 0.0051 & 0.0736 \\
    $F{\le}5\times$ $\uparrow$ & 0.7326 & \textbf{0.8116} & 0.2345 & 0.0105 & 0.2464 & 0.0451 & 0.1960 \\
    $F{\le}10\times$ $\uparrow$ & 0.8048 & \textbf{0.8536} & 0.3249 & 0.0208 & 0.3745 & 0.1209 & 0.2848 \\
    \bottomrule
  \end{tabular}
\end{table}

For $\sigma_p$, UniSTEM has a $10.2\%$ lower log-RMSE than the task expert (1.2285 versus 1.3676), a higher $R^2$ (0.6636 versus 0.5913), and a higher Pearson correlation (0.8356 versus 0.8064). For $\sigma_n$, its log-MAE is $12.8\%$ lower (0.6146 versus 0.7046), and its $R^2$ is 0.6082 versus 0.5278. The error in the difference between the two log-conductivities is also lower (0.2153 versus 0.2481).

The multiplicative factor makes the log-domain errors easier to interpret. UniSTEM's median $\sigma_p$ error factor is 1.737, compared with 1.996 for the expert. Its fraction of predictions within a factor of two is 0.5727, compared with 0.5015 for the expert. The external models have substantially larger median factors and negative $\sigma_p$ $R^2$ values. SenseNova-U1.5-8B produces nearly constant $\sigma_p$ predictions, so its corresponding Pearson correlation is undefined.

ATOMIC is the most extreme case in \cref{tab:supp_u4}: its $\sigma_p$ predictions collapse to unit scale, with $51.6\%$ exactly zero and $99.7\%$ at most $1$, whereas the ground truth spans $10^{-2.65}$ to $10^{6.49}$. Its large log-domain errors therefore reflect a gross scale mismatch rather than near-miss regression. The median $\sigma_p$ error factor is $9.7\times10^{11}$, and only $0.20\%$ of its predictions fall within a factor of $1.5$. For the same reason its $\Delta{=}\log\sigma_p{-}\log\sigma_n$ MAE (0.3210) is the lowest among the external models, since both of its predictions are driven to a common scale. Its $\sigma_p$ log-MAE and Spearman coefficient are reported in \cref{tab:main_results}.

\subsubsection{G1 DPC Field Reconstruction}
\label{sec:supp_g1}

\Cref{tab:supp_g1} expands the main results with whole-map quality, direction-error distributions, normalized field-component errors, and a tighter atomic-column-neighborhood score. The image-dependent neighborhood and its SSIM calculation are defined in \cref{sec:supp_adaptive_g1}.

\begin{table}[!htbp]
  \centering
  \caption{Supplementary G1 DPC field-reconstruction metrics on the full test set (3{,}904 samples). Scores already reported in \cref{tab:main_results} are omitted. Component metrics (DPCx/DPCy/magnitude) use the normalized field. The column-neighborhood metric is defined in \cref{sec:supp_adaptive_g1}.}
  \label{tab:supp_g1}
  \small
  \setlength{\tabcolsep}{4pt}
  \begin{tabular}{@{}lcccc@{}}
    \toprule
    \textbf{Metric} & \textbf{Expert} & \textbf{UniSTEM} & \textbf{BAGEL-7B} & \textbf{SenseNova-U1.5-8B} \\
    \midrule
    \textit{Whole-map} & & & & \\
    DPCDir PSNR (dB) $\uparrow$ & \textbf{26.66} & 25.63 & 6.46 & 6.03 \\
    \midrule
    \textit{Direction} & & & & \\
    Direction err.\ median ($^{\circ}$) $\downarrow$ & 7.12 & 7.12 & 90.38 & 114.38 \\
    Direction err.\ P90 ($^{\circ}$) $\downarrow$ & 32.38 & \textbf{29.63} & 163.88 & 179.88 \\
    Angle $\le10^{\circ}$ fraction $\uparrow$ & 0.6698 & 0.6694 & 0.0577 & 0.0475 \\
    Weighted angle MAE ($^{\circ}$) $\downarrow$ & \textbf{8.32} & 8.43 & 91.08 & 110.49 \\
    Direction MSE $\downarrow$ & \textbf{0.1135} & 0.1244 & 2.0064 & 1.9967 \\
    Direction $R^2$ $\uparrow$ & \textbf{0.8865} & 0.8756 & -1.0064 & -0.9967 \\
    \midrule
    \textit{Components (normalized, merged pixels)} & & & & \\
    DPCx MAE $\downarrow$ & \textbf{0.0353} & 0.0451 & 0.4172 & 0.5695 \\
    DPCx $R^2$ $\uparrow$ & \textbf{0.8963} & 0.8582 & -5.7712 & -12.4986 \\
    DPCy MAE $\downarrow$ & \textbf{0.0341} & 0.0433 & 0.3771 & 0.2895 \\
    DPCy $R^2$ $\uparrow$ & \textbf{0.8979} & 0.8581 & -4.9519 & -3.6158 \\
    Magnitude $M$ MAE $\downarrow$ & \textbf{0.0426} & 0.0578 & 0.4322 & 0.5559 \\
    Magnitude $M$ $R^2$ $\uparrow$ & \textbf{0.8763} & 0.8086 & -5.1959 & -10.2299 \\
    Vector MAE $\downarrow$ & \textbf{0.0546} & 0.0697 & 0.6181 & 0.6867 \\
    Vector RMSE $\downarrow$ & \textbf{0.0843} & 0.0990 & 0.6633 & 0.7959 \\
    \midrule
    \textit{Column-neighborhood} & & & & \\
    nb.\ SSIM (r3) $\uparrow$ & 0.8270 & 0.8377 & 0.0358 & 0.0224 \\
    \bottomrule
  \end{tabular}
\end{table}

At the tighter $3$\,px column neighborhood, UniSTEM reaches 0.8377 SSIM versus 0.8270 for the task expert, and its 90th-percentile direction error is $8.5\%$ lower (29.63\,$^\circ$ versus 32.38\,$^\circ$). The expert performs better on whole-map PSNR, weighted angular error, and the DPC component errors. These complementary measures show that the relative ordering depends on the image region and error type being evaluated.

\subsubsection{G2 Denoising}
\label{sec:supp_g2}

The main results table summarizes overall G2 denoising quality using the mean of per-image PSNR values. \Cref{tab:supp_g2} adds pixel errors, quality measures pooled across all test pixels, and errors stratified by input SNR. The pooled PSNR therefore has a different aggregation rule from the main-table PSNR. These complementary measures show whether the aggregate result is driven by a particular noise condition.

\begin{table}[!htbp]
  \centering
  \caption{Supplementary G2 denoising metrics on the full test set (16{,}048 samples). Scores already reported in \cref{tab:main_results} are omitted. External models are evaluated zero-shot. ``UNIT-denoise'' is a dedicated unsupervised STEM denoiser, and ``tk\_r\_em'' uses its released high-resolution STEM checkpoint~\cite{lobato2024deep}.}
  \label{tab:supp_g2}
  \small
  \setlength{\tabcolsep}{4pt}
  \begin{tabular}{@{}lcccccc@{}}
    \toprule
    \textbf{Metric} & \textbf{Expert} & \textbf{UniSTEM} & \textbf{BAGEL-7B} & \textbf{UNIT-denoise} & \textbf{SenseNova-U1.5-8B} & \textbf{tk\_r\_em} \\
    \midrule
    MAE $\downarrow$ & 7.503 & 6.956 & 111.079 & 97.269 & 86.306 & 53.668 \\
    RMSE $\downarrow$ & 10.167 & 9.411 & 123.634 & 108.763 & 107.550 & 57.110 \\
    Pooled PSNR (dB) $\uparrow$ & 22.803 & 23.923 & 6.085 & 5.872 & 6.282 & 9.947 \\
    Pooled $R^2$ $\uparrow$ & 0.973 & 0.979 & -0.281 & -0.345 & -0.224 & 0.474 \\
    \midrule
    \multicolumn{7}{l}{\textit{MAE by SNR (mean $\downarrow$)}} \\
    SNR 20 & 4.297 & 3.849 & 109.601 & 96.080 & 58.665 & 29.989 \\
    SNR 10 & 5.668 & 4.921 & 110.534 & 96.485 & 79.206 & 43.317 \\
    SNR 5 & 7.553 & 6.661 & 110.797 & 97.281 & 97.125 & 57.662 \\
    SNR 2 & 12.493 & 12.395 & 113.385 & 99.231 & 110.230 & 83.704 \\
    \bottomrule
  \end{tabular}
\end{table}

On the full test set, UniSTEM has a $7.3\%$ lower MAE than the task expert (6.956 versus 7.503) and a $7.4\%$ lower RMSE (9.411 versus 10.167). Its pooled PSNR and $R^2$ are also higher. As the input SNR falls from 20 to 2, both models' errors increase. UniSTEM retains a lower MAE in all four SNR groups, although the two models are close at the strongest noise level, SNR 2 (12.395 versus 12.493). Thus, the aggregate advantage extends across the evaluated noise conditions, with little separation at the lowest SNR.

Among the external models, \texttt{tk\_r\_em} generally has the best supplementary error scores, but it remains well behind UniSTEM and the task expert. The SNR-stratified results show how the models' errors change with noise strength beyond a single aggregate score.

\subsection{Evaluation on Real Electron-Microscopy Images}
\label{sec:supp_real_denoising}

We compare UniSTEM with a non-learned adaptive Wiener filter and two dedicated microscopy denoisers on three experimentally acquired HAADF images. The Wiener filter operates directly on native 16-bit intensities. For each pixel, it computes the local mean $\mu$ and variance $v$ in a $5\times5$ window, estimates the noise variance $\hat{\sigma}_n^2$ as the mean of the local variances over the image, and maps input intensity $x$ to
\begin{equation}
\hat{x}=\mu+\max\!\left(0,1-\frac{\hat{\sigma}_n^2}{v}\right)(x-\mu).
\end{equation}
Thus, when the local variance approaches the estimated noise variance, the output approaches the local mean. For UNIT-denoise, we use the authors' public pretrained UNIT-for-STEM-denoise model~\cite{wu2026unsupervised}. For \texttt{tk\_r\_em}, we use its high-resolution STEM checkpoint~\cite{lobato2024deep}.

These experimental images have no noise-free reference, so reference-based PSNR and SSIM cannot be evaluated~\cite{wang2020noise2atom}. Instead, we compute two diagnostics on the same prespecified lattice region for all outputs of each image. \emph{Raw-image correlation} ($r$) is the Pearson correlation between the raw image and the denoised output after both are Gaussian-smoothed with $\sigma=1.5$\,pixels. It measures consistency with the observation. For \emph{high-frequency reduction} ($\Delta HF$), we first fit an affine map from each output to the 16-bit raw-intensity scale using the smoothed images, then measure the decrease in median Fourier power spectral density over $0.30$--$0.45$ cycles/pixel, in dB. A larger $\Delta HF$ does not necessarily indicate better denoising, since blurring can also suppress high frequencies. The fractional (left, top, right, bottom) region bounds for the top, middle, and bottom rows are $(0.15,0.15,0.85,0.85)$, $(0.55,0.08,0.95,0.45)$, and $(0.30,0.30,0.70,0.70)$, respectively.

In the enlarged regions of \cref{fig:supp_real_denoising}, Wiener filtering reduces grainy noise but visibly softens the atomic-column edges, especially in the third image. UNIT-denoise retains pronounced graininess in the first two images and produces numerous spiky artifacts around atomic columns in the third. The lattices produced by \texttt{tk\_r\_em} appear smoother and more regular. UniSTEM retains relatively distinct atomic columns, although some background texture remains in the second and third images. These visual differences should be considered alongside the diagnostics rather than ranking outputs by $\Delta HF$ alone.

Wiener filtering has the highest $r$ in all three images ($0.998/0.995/0.987$), but lower $\Delta HF$ than every learned method and visibly blurrier outputs. Among learned methods, \texttt{tk\_r\_em} has the highest $r$ ($0.992/0.966/0.898$), followed closely by UniSTEM ($0.990/0.954/0.873$). Both exceed UNIT-denoise. \texttt{tk\_r\_em} has greater high-frequency reduction in the first two images, whereas UniSTEM has greater reduction in the third ($46.82$ versus $42.67$\,dB). UniSTEM was trained only on simulated micrographs, with none of these real images used for training. Its results therefore remain comparable to those of dedicated denoisers under simulation-to-real transfer, although no method can be identified as closest to an unknown noise-free image from these diagnostics alone.

\noindent\begin{minipage}{\textwidth}
  \centering
  \includegraphics[width=\textwidth]{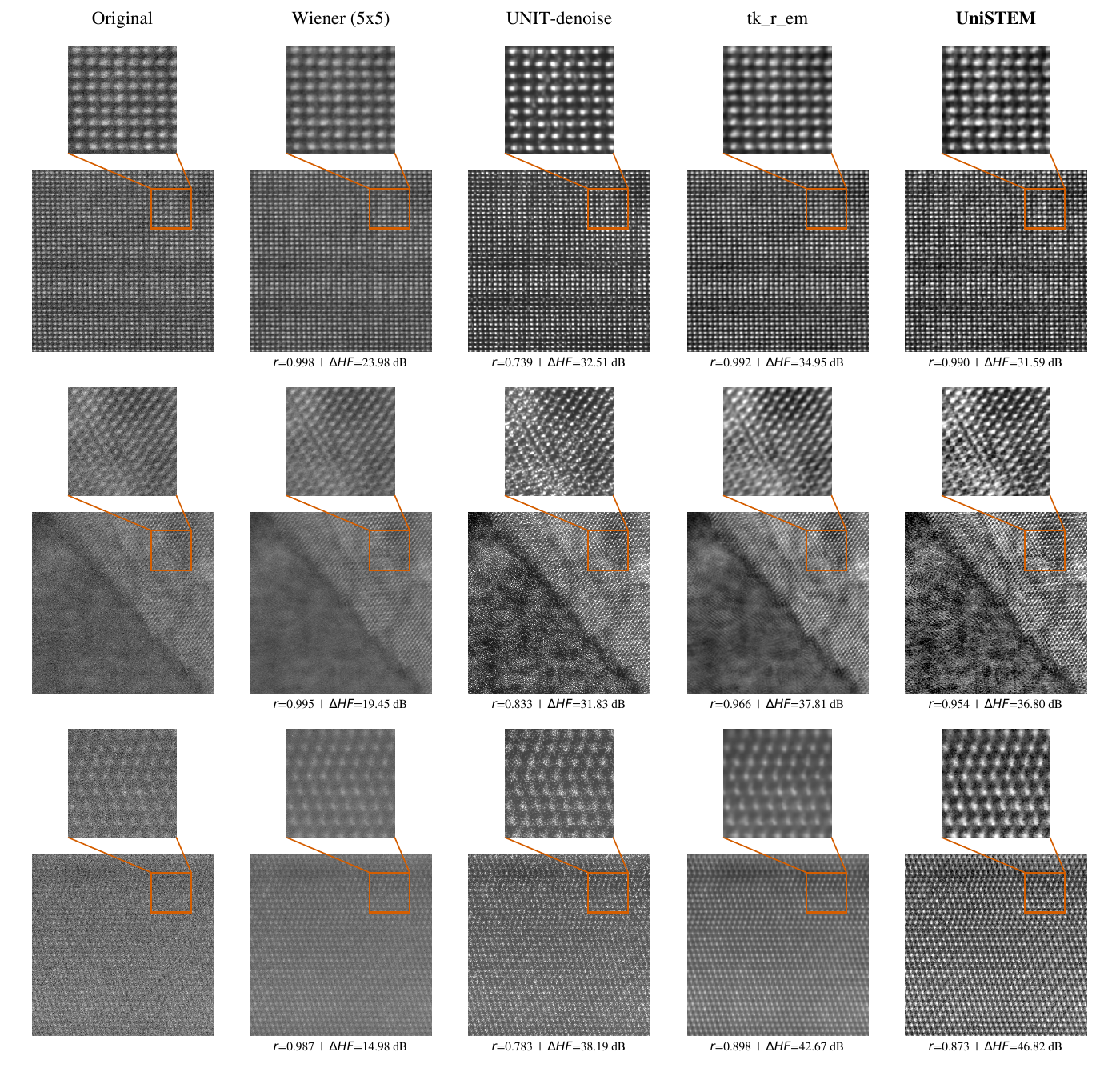}
  \captionof{figure}{Denoising of three real HAADF images. Columns show the raw observation, a non-learned $5\times5$ adaptive Wiener filter, UNIT-denoise, \texttt{tk\_r\_em}, and UniSTEM. In each panel, the upper image enlarges the orange-boxed region of the overview below. Connectors identify the crop, whose fractional bounds are $(x_0,y_0,x_1,y_1)=(0.66,0.10,0.88,0.32)$ for every method in a row. Annotations below the overviews give raw-image correlation $r$ and high-frequency reduction $\Delta HF$ on the same prespecified lattice region for each row. Rows represent distinct examples without assigning material names to acquisition identifiers.}
  \label{fig:supp_real_denoising}
\end{minipage}

\subsection{Evaluation on Other Benchmarks}
\label{sec:supp_temi_denoising}

To test whether G2 denoising transfers beyond our own benchmark, we evaluate on the independently released TEMImageNet v1.3 and its Multislice variant~\cite{lin2021atomsegnet}. These datasets provide simulated ADF-STEM micrographs and noise-free targets at $256\times256$ resolution. We stratify-sample 1{,}200 images from each repository and exclude pairs whose noisy observation is identical to its target, leaving 2{,}117 valid pairs. All metrics follow \cref{tab:supp_g2} and are computed in the 8-bit $[0,255]$ intensity domain. This evaluation probes transfer outside our training simulation pipeline.

We evaluate UniSTEM and four dedicated denoisers zero-shot, without further training or per-image calibration on TEMImageNet. The baselines use their released checkpoints: the high-resolution STEM checkpoint of \texttt{tk\_r\_em}~\cite{lobato2024deep}, UNIT-denoise~\cite{wu2026unsupervised}, the HAADF enhancement checkpoint of SFIN~\cite{li2025noise}, and SCGN~\cite{SCGN}. Inputs follow the format required by each model's released code.

\begin{table}[!htbp]
  \centering
  \caption{Cross-simulator denoising on TEMImageNet (2{,}117 valid pairs, zero-shot). The ``Noisy input'' column reports the PSNR of the unprocessed observation against the noise-free target. All models are evaluated without further training on TEMImageNet.}
  \label{tab:supp_temi_denoise}
  \small
  \setlength{\tabcolsep}{3.5pt}
  \begin{tabular}{@{}lcccccc@{}}
    \toprule
    \textbf{Metric} & \textbf{Noisy input} & \textbf{UniSTEM} & \textbf{tk\_r\_em} & \textbf{UNIT-denoise} & \textbf{SFIN} & \textbf{SCGN} \\
    \midrule
    PSNR (dB) $\uparrow$ & 16.357 & \textbf{23.500} & 17.716 & 15.308 & 13.863 & 10.328 \\
    PSNR gain (dB) $\uparrow$ & --- & \textbf{+7.143} & +1.358 & $-$1.050 & $-$2.494 & $-$6.029 \\
    SSIM $\uparrow$ & --- & 0.7725 & \textbf{0.7882} & 0.4267 & 0.4408 & 0.1488 \\
    MAE $\downarrow$ & --- & \textbf{19.069} & 26.844 & 36.871 & 47.252 & 62.844 \\
    RMSE $\downarrow$ & --- & \textbf{22.775} & 36.398 & 47.285 & 56.867 & 82.080 \\
    $R^2$ $\uparrow$ & --- & \textbf{0.598} & 0.434 & $-$0.109 & $-$0.654 & $-$2.153 \\
    Pearson $r$ $\uparrow$ & --- & 0.956 & \textbf{0.994} & 0.822 & 0.952 & 0.049 \\
    \bottomrule
  \end{tabular}
\end{table}

\begin{figure}[!htbp]
  \centering
  \includegraphics[width=\textwidth]{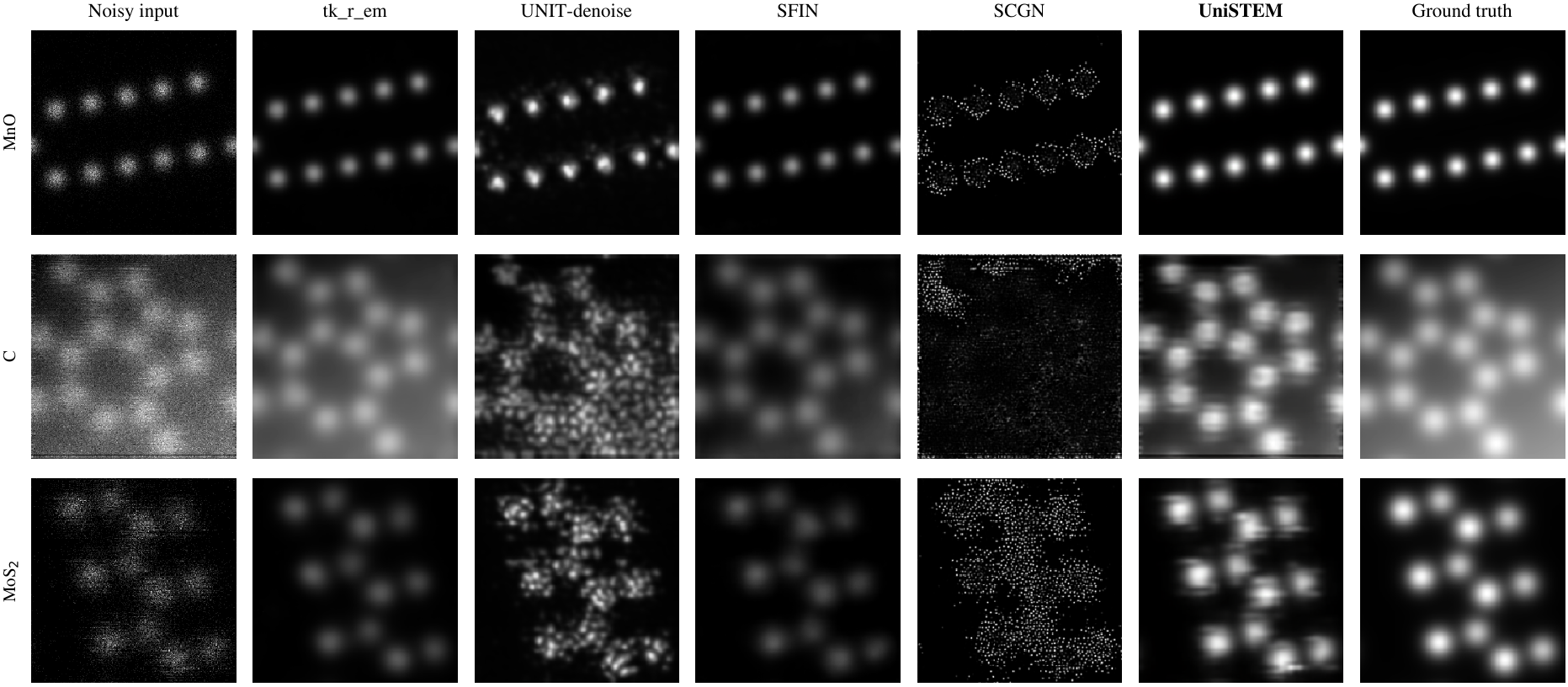}
  \caption{Qualitative denoising comparisons on three TEMImageNet examples. The first two rows show MnO and C from TEMImageNet v1.3. The third shows MoS$_2$ from its Multislice variant. Columns show the noisy input, four dedicated denoisers (\texttt{tk\_r\_em}, UNIT-denoise, SFIN, and SCGN), UniSTEM, and the noise-free ground truth.}
  \label{fig:supp_temi_denoise}
\end{figure}

As shown in \cref{tab:supp_temi_denoise}, UniSTEM reaches $23.500$\,dB PSNR, a gain of $7.143$\,dB over the noisy input and $5.784$\,dB over the strongest baseline, \texttt{tk\_r\_em}. It also has the lowest MAE and RMSE and the highest $R^2$. The SSIM and Pearson correlation of \texttt{tk\_r\_em} ($0.7882$ and $0.994$) slightly exceed those of UniSTEM ($0.7725$ and $0.956$), so UniSTEM does not lead on every structural metric. The other three released baselines have lower PSNR than the noisy input on this external set.

\Cref{fig:supp_temi_denoise} compares three outputs with their noise-free targets. UniSTEM restores clear atomic columns in MnO and the main lattice patterns in the noisier C and MoS$_2$ examples, although local intensity and texture still differ from the targets. UNIT-denoise exhibits fragmented textures, while SCGN retains only sparse bright points. SFIN and \texttt{tk\_r\_em} produce smoother images, but their intensities and details differ from the targets. Together, these results show that UniSTEM's G2 denoising ability extends beyond our own benchmark to independently released simulated micrographs under the same zero-shot evaluation protocol for all models.

\begin{figure}[!htbp]
  \centering
  \begin{subfigure}[b]{0.22\linewidth}
    \centering
    \includegraphics[height=0.20\textheight,width=\linewidth,keepaspectratio]{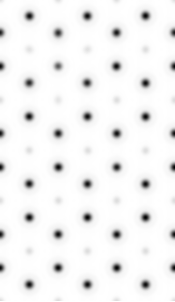}
    \caption{Input (BF, \#06051)}
  \end{subfigure}\hfill
  \begin{subfigure}[b]{0.22\linewidth}
    \centering
    \includegraphics[height=0.20\textheight,width=\linewidth,keepaspectratio]{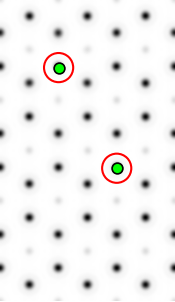}
    \caption{GT vacancies + adaptive radius}
  \end{subfigure}\hfill
  \begin{subfigure}[b]{0.22\linewidth}
    \centering
    \includegraphics[height=0.20\textheight,width=\linewidth,keepaspectratio]{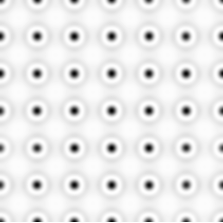}
    \caption{Input (ABF, \#00505)}
  \end{subfigure}\hfill
  \begin{subfigure}[b]{0.22\linewidth}
    \centering
    \includegraphics[height=0.20\textheight,width=\linewidth,keepaspectratio]{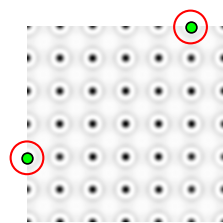}
    \caption{GT vacancies + adaptive radius}
  \end{subfigure}
  \caption{Image-dependent vacancy-matching radius for two U1 test images. The disks around ground-truth vacancies use $R{=}0.5\,d$, with $d$ estimated independently from each input micrograph.}
  \label{fig:supp_adaptive_u1}
\end{figure}

\subsection{Adaptive Metric Definitions and Rationale}
\label{sec:supp_adaptive}

The two image-adaptive indicators used in the main results are defined below. U1 scales its matching radius with a spacing estimated from the image spectrum. G1 uses detected atomic-column centers and scales the neighborhood radius with their median nearest-neighbor spacing.

\subsubsection{U1 Adaptive-Radius Vacancy \texorpdfstring{$F_1$}{F1}}
\label{sec:supp_adaptive_u1}

For each test micrograph, we estimate one \emph{global} column spacing from the centered magnitude spectrum of its grayscale image. Let $\mathbf{f}^{*}=(f_x^{*},f_y^{*})$ be the strongest non-DC peak after excluding a central disk of radius two frequency samples. The spacing in pixels is
\begin{equation}
  d = 1\,\Big/\!\sqrt{(f_x^{*})^{2}+(f_y^{*})^{2}}.
\end{equation}
We retain estimates in the range $2\le d\le \tfrac{1}{2}\min(H,W)$ and skip images without a valid peak. The matching radius is $R=k\,d$ for $k\in\{0.25,0.35,0.5\}$. Predicted and ground-truth vacancies are paired greedily in ascending order of distance. A pair is a true positive if its distance is at most $R$ and neither point has already been matched. Unmatched predictions and ground-truth vacancies are false positives and false negatives, respectively. Precision, recall, and $F_1$ are computed from counts pooled over the evaluated images.

A fixed pixel radius represents different fractions of the lattice period across materials. Scaling the radius with $d$ makes the criterion comparable across those periods. For a regular lattice with nearest-neighbor spacing $d$, $k{=}0.5$ reaches the midpoint between neighboring columns, while smaller values require tighter localization. This interpretation is approximate for images containing multiple column sublattices. In HAADF images, bright heavy-element columns can dominate the spectral peak and bias $d$ toward their sublattice. BF and ABF images may yield another dominant period. Every model is evaluated with the same spacing estimate for a given image. \Cref{fig:supp_adaptive_u1} illustrates the resulting radii on BF and ABF examples.

\subsubsection{G1 Column-Neighborhood SSIM}
\label{sec:supp_adaptive_g1}

Whole-map SSIM includes extensive background regions. The column-neighborhood metric instead averages local SSIM over an image-dependent mask centered on atomic columns. Both the selected centers and the neighborhood radius adapt to each image.

\begin{figure}[!htbp]
  \centering
  \begin{minipage}{0.82\textwidth}
  \centering
  \begin{subfigure}[b]{0.31\linewidth}
    \centering
    \includegraphics[width=\linewidth]{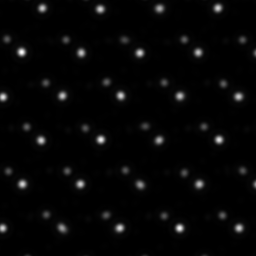}
    \caption{HAADF input (\#10000)}
  \end{subfigure}\hfill
  \begin{subfigure}[b]{0.31\linewidth}
    \centering
    \includegraphics[width=\linewidth]{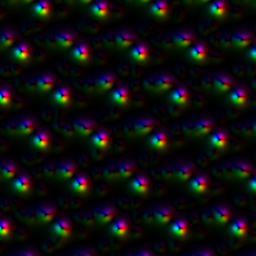}
    \caption{GT DPCDir field}
  \end{subfigure}\hfill
  \begin{subfigure}[b]{0.31\linewidth}
    \centering
    \includegraphics[width=\linewidth]{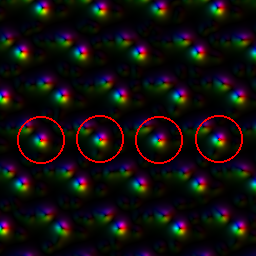}
    \caption{DPCDir + nb.\ SSIM regions}
  \end{subfigure}
  \\[1ex]
  \begin{subfigure}[b]{0.31\linewidth}
    \centering
    \includegraphics[width=\linewidth]{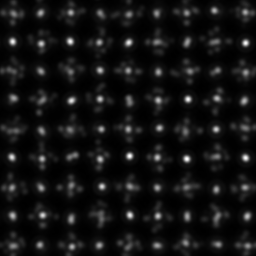}
    \caption{HAADF input (\#10050)}
  \end{subfigure}\hfill
  \begin{subfigure}[b]{0.31\linewidth}
    \centering
    \includegraphics[width=\linewidth]{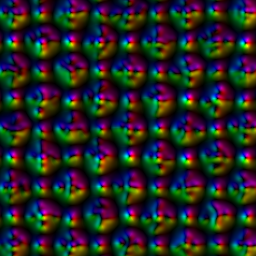}
    \caption{GT DPCDir field}
  \end{subfigure}\hfill
  \begin{subfigure}[b]{0.31\linewidth}
    \centering
    \includegraphics[width=\linewidth]{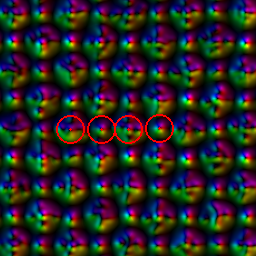}
    \caption{DPCDir + nb.\ SSIM regions}
  \end{subfigure}
  \end{minipage}
  \caption{Column-neighborhood SSIM regions for G1. For two test samples, we show the HAADF input, the ground-truth DPCDir field, and the same field annotated with four representative red circles. The centers are detected from the HAADF input, and each image uses an adaptive radius $R=0.5d$, where $d$ is its median nearest-neighbor column spacing. The circles illustrate the column neighborhoods on the corresponding DPC field.}
  \label{fig:supp_adaptive_g1}
\end{figure}

Atomic-column centers are detected from the aligned HAADF image (which is registered to the DPCDir field). The image is smoothed with a Gaussian of $\sigma{=}1$\,px, and a pixel is retained as a center if it is a local maximum over its $11{\times}11$ neighborhood and its intensity exceeds $45\%$ of the brightest peak,
\begin{equation}
  \mathcal{C}=\bigl\{\,c:\ \tilde{I}(c)=\max_{\mathcal{N}_{11}(c)}\tilde{I},\ \ \tilde{I}(c)>0.45\,\tilde{I}_{\max}\,\bigr\}.
\end{equation}
The per-pixel SSIM between the predicted and ground-truth DPCDir maps is computed with an $11{\times}11$ Gaussian window ($\sigma{=}1.5$, dynamic range $L{=}255$),
\begin{equation}
  \begin{aligned}
    \mathrm{SSIM}(x)&=\frac{2\mu_x\mu_y+C_1}{\mu_x^{2}+\mu_y^{2}+C_1}
      \frac{2\sigma_{xy}+C_2}{\sigma_x^{2}+\sigma_y^{2}+C_2},\\
    C_1&=(0.01L)^{2},\qquad C_2=(0.03L)^{2}.
  \end{aligned}
\end{equation}
Let $d$ be the median nearest-neighbor distance among the detected centers in $\mathcal{C}$. The primary neighborhood radius is $R=0.5d$. Neighborhood SSIM is the mean of $\mathrm{SSIM}(x)$ over the union of radius-$R$ disks centered at the detected columns,
\begin{equation}
  \begin{aligned}
    \mathcal{M}_R&=\bigcup_{c\in\mathcal{C}}\{\,x:\|x-c\|\le R\,\},\\
    \mathrm{nb.\ SSIM}(R)&=\frac{1}{|\mathcal{M}_R|}
      \sum_{x\in\mathcal{M}_R}\mathrm{SSIM}(x).
  \end{aligned}
\end{equation}
The fixed $r{=}3$\,px score in \cref{tab:supp_g1} provides a tighter auxiliary comparison. The metric averages SSIM only over the selected mask rather than subtracting a background score. The $45\%$ intensity threshold preferentially retains bright columns in HAADF images, so the reported score reflects reconstruction quality near those selected columns rather than a uniform average over every atomic species. \Cref{fig:supp_adaptive_g1} illustrates four adaptive-radius neighborhoods per image.

\subsection{Error Coupling and Condition Robustness}

\begin{table}[!htbp]
  \centering
  \caption{U4--G1 error coupling on 3{,}904 shared samples. $\rho$ and $\varphi$ measure association. Q1--Q3 group samples by G1 error.}
  \label{tab:supp_coupling_sample}
  \small
  \setlength{\tabcolsep}{4pt}
  \begin{tabular}{@{}lccccc@{}}
    \toprule
    \textbf{Model} & $\rho$ [95\% CI] & $\varphi$ [95\% CI] & @Q1 & @Q2 & @Q3 \\
    \midrule
    UniSTEM & $-0.003$ [$-0.035$, $0.027$] & $0.014$ [$-0.015$, $0.044$] & 0.676 & 0.625 & 0.649 \\
    Field expert & $-0.007$ [$-0.043$, $0.022$] & $0.027$ [$-0.002$, $0.057$] & 0.808 & 0.738 & 0.769 \\
    \bottomrule
  \end{tabular}
\end{table}

\begin{figure}[!htbp]
  \centering
  \includegraphics[width=0.62\textwidth]{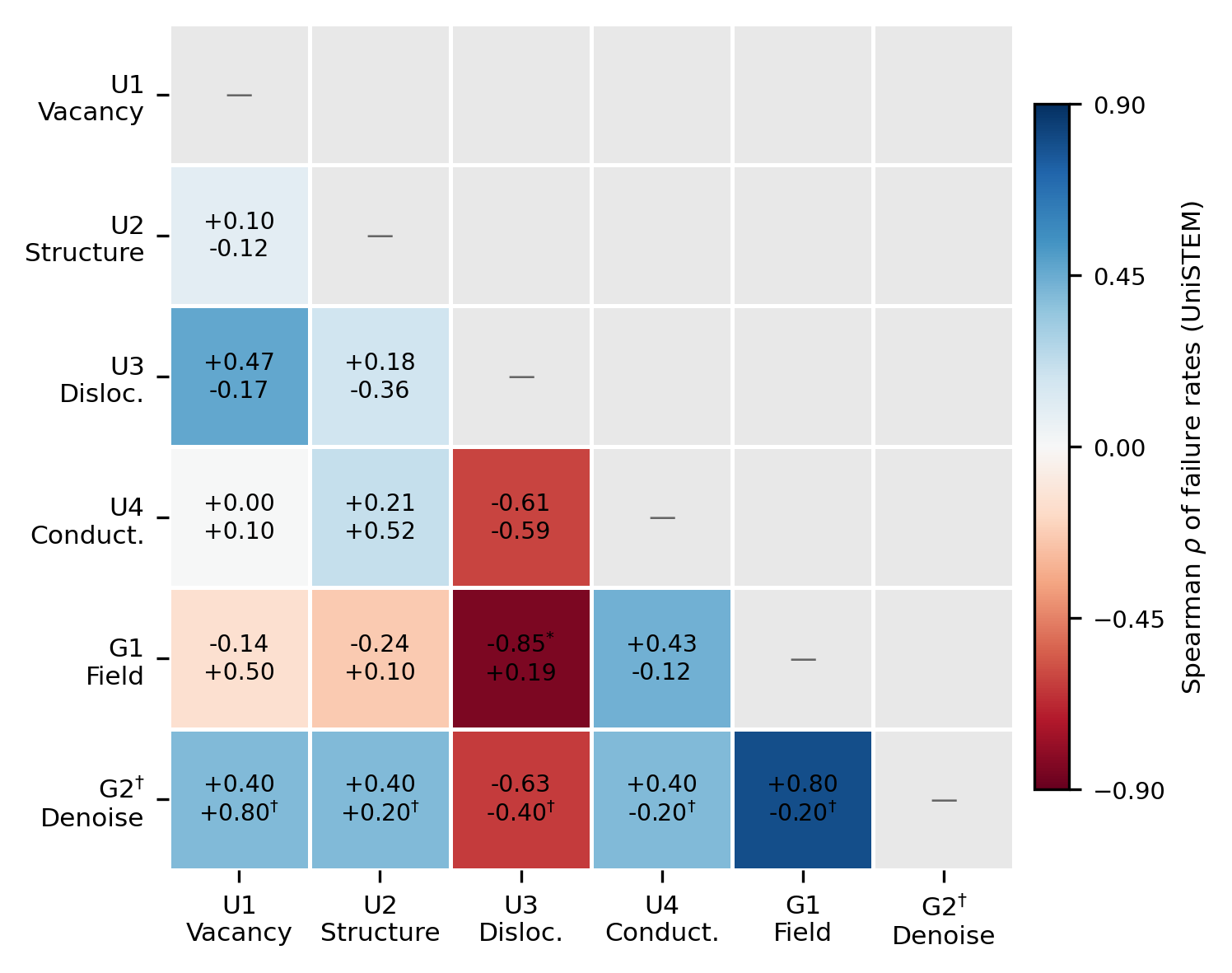}
  \caption{Correlations of task failure rates across imaging-condition bins. Each cell shows UniSTEM/expert Spearman $\rho$. $^{*}$ marks uncorrected $p<0.05$ and $^{\dagger}$ marks G2 pairs with only detector-mode bins.}
  \label{fig:coupling_heatmap}
\end{figure}

\begin{table}[t]
  \centering
  \caption{Win--tie--loss counts against task experts across imaging-condition slices. A tie means a difference within 1\%. Dashes mark unavailable axes.}
  \label{tab:supp_slice_wtl}
  \small
  \setlength{\tabcolsep}{2.5pt}
  \begin{tabular}{@{}lcccccc@{}}
    \toprule
    \textbf{Axis (slices)} & \textbf{U1} & \textbf{U2} & \textbf{U3} & \textbf{U4} & \textbf{G1} & \textbf{G2} \\
    \midrule
    Beam energy (2)   & 0--1--1 & 0--1--1 & 2--0--0 & 2--0--0 & 2--0--0 & - \\
    Detector mode (4) & 0--0--4 & 0--2--2 & 4--0--0 & 4--0--0 & 3--1--0 & 3--0--1 \\
    Defocus (2)       & 0--0--2 & 1--0--1 & 1--1--0 & 2--0--0 & 2--0--0 & - \\
    Zone axis (3)     & - & 0--2--1 & - & - & - & - \\
    Elements (3)      & - & 1--2--0 & - & 3--0--0 & 3--0--0 & - \\
    SNR (4)           & - & - & - & - & - & 3--1--0 \\
    \bottomrule
  \end{tabular}
\end{table}

\begin{table}[!htbp]
  \centering
  \caption{UniSTEM/expert values on each task's primary metric across imaging conditions. Dashes mark unavailable axes.}
  \label{tab:supp_slice_delta}
  \footnotesize
  \setlength{\tabcolsep}{4pt}
  \begin{tabular}{@{}llcccccc@{}}
    \toprule
    \textbf{Axis} & \textbf{Slice} & \textbf{U1} & \textbf{U2} & \textbf{U3} & \textbf{U4} & \textbf{G1} & \textbf{G2} \\
    \midrule
    \multirow{2}{*}{Energy} & 200 kV & .322/.421 & .562/.568 & .768/.730 & .698/.796 & 15.16/16.28 & - \\
                            & 300 kV & .420/.419 & .560/.563 & .685/.631 & .591/.741 & 15.23/18.85 & - \\
    \midrule
    \multirow{4}{*}{Mode} & ABF   & .386/.403 & .559/.565 & .738/.676 & .655/.764 & 17.32/17.44 & 27.50/27.19 \\
                          & BF    & .404/.417 & .559/.568 & .715/.657 & .647/.770 & 15.85/16.93 & 30.20/30.58 \\
                          & ADF   & .388/.491 & .563/.558 & .709/.686 & .654/.776 & 12.45/16.25 & 36.68/35.42 \\
                          & HAADF & .319/.370 & .563/.571 & .718/.680 & .644/.775 & 15.17/19.15 & 37.74/36.71 \\
    \midrule
    \multirow{2}{*}{Defocus} & neg    & .339/.409 & .556/.582 & .591/.585 & .611/.702 & 14.16/16.39 & - \\
                             & near 0 & .435/.440 & .564/.557 & .828/.748 & .691/.844 & 16.33/18.59 & - \\
    \midrule
    \multirow{3}{*}{Zone} & 001     & - & .524/.528 & - & - & - & - \\
                          & 110/111 & - & .702/.702 & - & - & - & - \\
                          & 210/211 & - & .606/.617 & - & - & - & - \\
    \midrule
    \multirow{3}{*}{Elements} & single & - & .970/.960 & - & .403/1.033 & 16.26/20.00 & - \\
                              & binary & - & .581/.586 & - & .571/.726 & 13.68/16.86 & - \\
                              & ternary+ & - & .538/.542 & - & .687/.785 & 15.78/17.63 & - \\
    \midrule
    \multirow{4}{*}{SNR} & 2  & - & - & - & - & - & 27.89/27.80 \\
                         & 5  & - & - & - & - & - & 32.26/31.38 \\
                         & 10 & - & - & - & - & - & 34.90/34.08 \\
                         & 20 & - & - & - & - & - & 37.08/36.64 \\
    \bottomrule
  \end{tabular}
\end{table}

\begin{figure}[!htbp]
  \centering
  \includegraphics[width=0.86\textwidth]{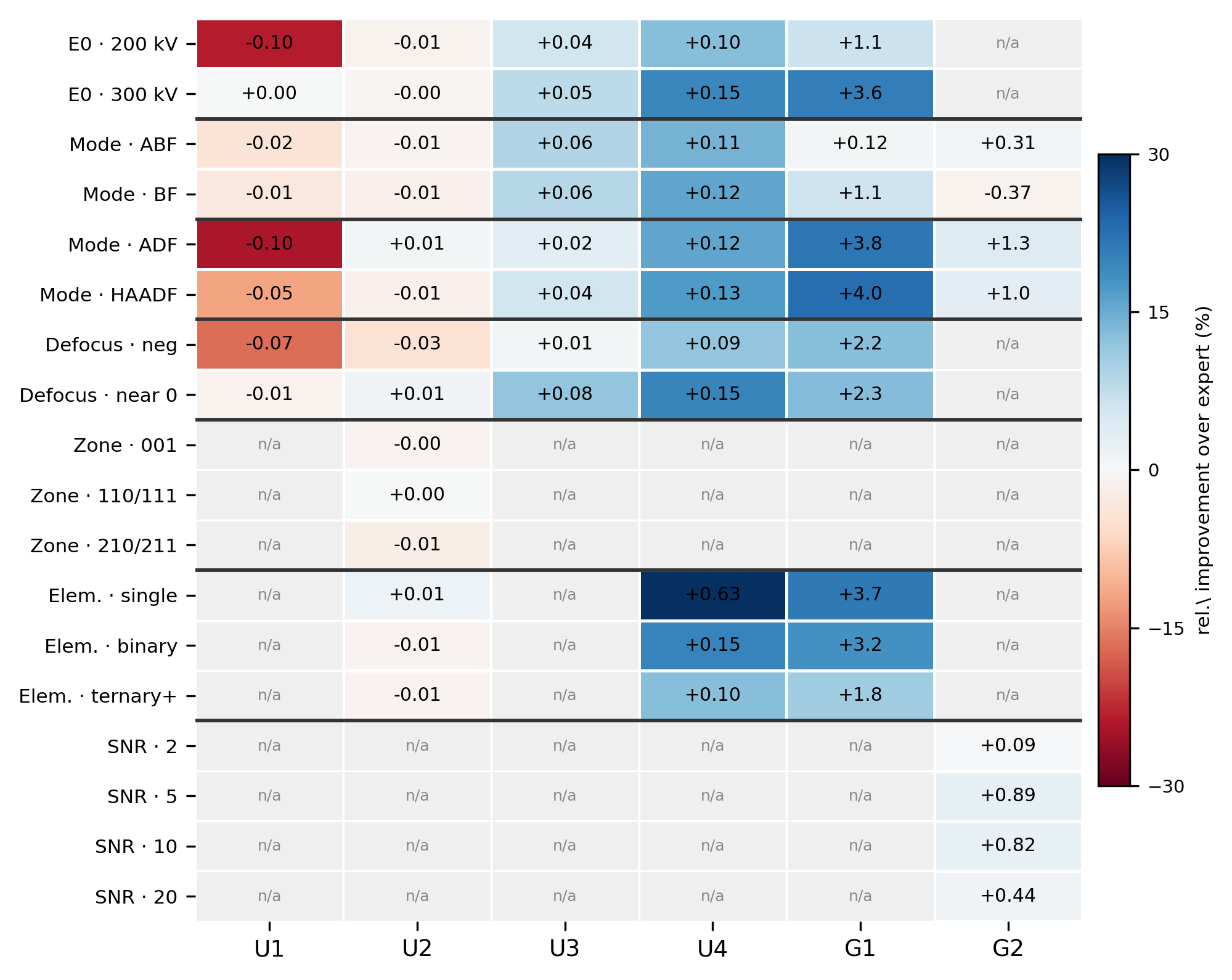}
  \caption{UniSTEM--expert performance gap across imaging conditions. Positive values favor UniSTEM. Gray cells indicate unavailable condition axes.}
  \label{fig:slice_delta}
\end{figure}

\label{sec:supp_coupling}
\label{sec:coupling_analysis}

Shared parameters may cause tasks to fail under the same images or imaging conditions, and full-test-set averages may conceal condition-specific deficits. We therefore examine per-sample error associations, associations between condition-level failure rates, and performance within each condition slice (\cref{fig:coupling_heatmap,fig:slice_delta}). These analyses measure association rather than causal error propagation.

All per-sample records come from the same full-test-set evaluations as \cref{tab:main_results}, and recomputing their aggregates reproduces the main-table values. We form slices from the corpus metadata for detector mode (ABF/BF/ADF/HAADF), beam energy (200/300\,kV), and defocus (neg/near0). We additionally examine zone axis for U2, element count for U2 and U4/G1, and input SNR for G2. Slices with fewer than 30 samples are omitted, and the smallest reported slice has $n=52$. Two original U3 samples lacked imaging conditions, corresponding to eight images across four detector modes. The beam-energy and defocus slices therefore use 292 of the 300 U3 images, while the detector-mode slices use all 300. We use the primary metric for each task in \cref{tab:main_results}. U1 and U3 slice F1 scores are pooled over samples, and G1 slice averages are weighted by pixel count.

Only U4 and G1 share the same 3{,}904 dual-annotated samples, allowing a direct test of co-failure on individual images. Their per-sample errors have little association for UniSTEM (Spearman $\rho=-0.003$, 95\% CI $[-0.035,\,0.027]$, $\varphi=0.014$) or the dual-output expert ($\rho=-0.007$, $\varphi=0.027$). Dividing samples into terciles of increasing G1 error leaves UniSTEM's mean U4 error nearly flat at $0.676/0.625/0.649$ (\cref{tab:supp_coupling_sample}). We find no evidence of substantial sample-level co-failure for this task pair.

Other task pairs use different test images, so \cref{fig:coupling_heatmap} compares failure rates across shared imaging-condition bins rather than co-failure on the same image. A sample fails when its error exceeds the median error of the corresponding task expert. Four detector modes, two beam-energy groups, and two defocus groups provide eight bins, while pairs involving G2 share only the four detector-mode bins. The dark-red U3--G1 cell shows a strong negative correlation ($\rho=-0.85$), indicating opposite rankings of difficult conditions. Yet the tasks use different images, and the correlation uses only eight bins. U3 also has just 72--76 images per detector mode compared with 976 for G1, making its bin failure rates more sensitive to individual samples. Its uncorrected $p=0.009$ does not remain significant after correction across 15 task pairs, so the color does not establish a task trade-off. The dark-blue G1--G2 cell shows a positive correlation ($\rho=0.80$) across detector modes. Although each mode contains 976 G1 and 4{,}012 G2 images, this correlation is based on only four bins and is also not significant after correction. The strong colors therefore do not by themselves establish stable cross-task error coupling.

\Cref{fig:slice_delta,tab:supp_slice_wtl,tab:supp_slice_delta} compares UniSTEM with each task expert within condition slices. UniSTEM wins all 11 U4 slices, 10 of 11 G1 slices, 7 of 8 U3 slices, and 6 of 8 G2 slices. G1 reduces direction-angle error by about $3.6^{\circ}$ at 300\,kV and $4.0^{\circ}$ in HAADF, where its expert is weaker. U2 has 2 wins, 7 ties, and 5 losses across 14 slices, remaining close to its expert overall. Under the fixed 5\,px tolerance, U1 is weaker at 200\,kV and in ADF, with F1 scores $23.5\%$ and $21.0\%$ below its expert, respectively (absolute gaps of $-0.099$ and $-0.103$). It is close to its expert at 300\,kV ($0.420$ versus $0.419$). G2 is also slightly below its expert in BF ($30.20$ versus $30.58$\,dB). Overall, the gains for most tasks span several conditions, while specific imaging settings remain weaker.

\begin{table}[!htbp]
  \centering
  \caption{Gradient alignment between U1 and the other understanding tasks at five checkpoints. The U1 task-expert checkpoint precedes mixed-task training and serves as the initialization reference. Positive cosine indicates aligned update directions. Cross/self measures other tasks' contribution to the U1 direction. $\|g_{U1}\|$ is the norm of U1's own gradient (arbitrary units, averaged over the 30 probe batches) and shrinks by more than an order of magnitude along the lineage.}
  \label{tab:supp_grad_probe}
  \small
  \setlength{\tabcolsep}{6pt}
  \begin{tabular}{@{}lcccccc@{}}
    \toprule
    \textbf{Checkpoint} & $\cos(g_{U1},g_{U2})$ & $\cos(g_{U1},g_{U3})$ & $\cos(g_{U1},g_{U4})$ & cross/self & total$/\|\bar g\|$ & $\|g_{U1}\|$ \\
    \midrule
    Task expert      & $-0.044$ & $-0.059$ & $-0.077$ & $-1.27$ & $-0.016$ & $0.98$  \\
    S1 (EMA weights)           & $+0.321$ & $+0.570$ & $+0.130$ & $1.08$  & $0.741$  & $0.35$  \\
    S1 (raw weights)           & $+0.024$ & $+0.019$ & $+0.034$ & $0.16$  & $0.291$  & $0.18$  \\
    S2 (MOPD, b24 reverse-KL)  & $+0.081$ & $+0.034$ & $+0.094$ & $1.15$  & $0.202$  & $0.080$ \\
    S2 + B1 replay             & $+0.051$ & $+0.021$ & $-0.003$ & $0.20$  & $0.108$  & $0.054$ \\
    \bottomrule
  \end{tabular}
\end{table}

\section{Training Dynamics of the Understanding Tower}

\subsection{Gradient Alignment among Understanding Tasks}
\label{sec:supp_grad_probe}

\Cref{fig:stage_progression} shows that U1 adaptive F1 falls and later recovers along the training path. Because U1--U4 share the understanding trunk, we test whether this change is accompanied by persistently opposing task-gradient directions. At each checkpoint, we evaluate all four tasks at the same model weights and average the gradient of unweighted token-level cross-entropy over 30 mini-batches from each task's training stream. A disjoint set of 15 mini-batches per task is used for significance testing. The cosine between two task gradients measures the alignment of their local update directions.

Let $\bar g=\frac14\sum_{k=1}^{4}g_{U_k}$, $\hat g_{U1}=g_{U1}/\|g_{U1}\|$, and $c_k=\cos(g_{U_k},g_{U1})$. The projection of the equal-weight task gradient onto the U1 direction can be decomposed as
\begin{equation}
\begin{aligned}
P_{\mathrm{self}}  &= \tfrac14\|g_{U1}\|,\\
P_{\mathrm{cross}} &= \tfrac14\sum_{k=2}^{4}c_k\|g_{U_k}\|,\\
\langle\bar g,\hat g_{U1}\rangle
                  &= P_{\mathrm{self}}+P_{\mathrm{cross}}.
\end{aligned}
\end{equation}
The ratio $P_{\mathrm{cross}}/P_{\mathrm{self}}$ compares the other tasks' projection along U1 with U1's own term. \Cref{tab:supp_grad_probe} also reports $\langle\bar g,\hat g_{U1}\rangle/\|\bar g\|$ and the U1 gradient norm. Significance is assessed by a sign-flip permutation test on per-pack cosines with 5{,}000 permutations and a fixed seed.

Gradient alignment changes across checkpoints. At this pre-mixing U1 expert checkpoint, U1 is mildly anti-aligned with U2--U4 ($|\cos|\le0.08$). At this pre-mixing reference, the U3 and U4 entries have uncorrected $p<0.01$, while U2 has $p=0.10$. After mixed training, the S1 EMA weights show positive alignment, and the S1 raw-weight cosines remain positive but close to zero. At S2, the U1 cosines with U2, U3, and U4 are $0.081$, $0.034$, and $0.094$. After U1 replay they are $0.051$, $0.021$, and $-0.003$. The ratios $P_{\mathrm{cross}}/P_{\mathrm{self}}$ at S1 EMA and S2 are $1.08$ and $1.15$. Across the four post-mixing checkpoints, none shows statistically significant negative alignment with U1, and the other tasks have positive U1-directed gradient components at some stages.

This gradient pattern is consistent with the U1 evaluation results in \cref{tab:supp_u1}. Relative to the U1 expert, the final model scores $25.2\%$ lower at F1@2px ($0.184$ versus $0.246$) and $11.0\%$ lower at F1@5px ($0.374$ versus $0.420$). Its count error is $22.0\%$ lower ($0.128$ versus $0.164$). As the matching tolerance scales with atomic-column spacing, the gap narrows and UniSTEM leads at $k=0.35$ ($0.576$ versus $0.537$) and $k=0.5$ ($0.643$ versus $0.591$). The remaining U1 deficit is therefore concentrated in strict coordinate matching. Together with \cref{fig:stage_progression} and the sampling-ratio experiments in \cref{sec:supp_ratio}, the fall and later recovery of U1 align more closely with changes in training composition and U1 replay than with a stable directional conflict between task gradients.

\subsection{Sampling-Ratio Sensitivity of the Understanding Tower}
\label{sec:supp_ratio}

\begin{table}[!htbp]
  \centering
  \caption{Full-test-set results behind \cref{fig:supp_ratio_s11,fig:supp_ratio_s12,fig:supp_ratio_replay}. U1 is plotted using adaptive F1. Fixed F1@$5$\,px is included for comparison. U4 $\log_{10}$-MAE is lower when better, whereas all other metrics are higher when better.}
  \label{tab:supp_ratio}
  \small
  \setlength{\tabcolsep}{6pt}
  \begin{tabular}{@{}lcccccc@{}}
    \toprule
    \textbf{Level} & \textbf{U1 share} & \textbf{U1 F1@5px} & \textbf{U1 F1@adaptive} & \textbf{U2 formula} & \textbf{U3 F1@20px} & \textbf{U4 $\log_{10}$-MAE} \\
    \midrule
    \multicolumn{7}{l}{\emph{S1-1: mixed understanding SFT, U2--U4 weight $1.0$ each}} \\
    $w_{U1}{=}0.5$                & $14.3\%$ & $0.088$ & $0.192$ & $0.560$ & $0.707$ & $0.671$ \\
    $w_{U1}{=}1.0$                & $25.0\%$ & $0.167$ & $0.404$ & $0.535$ & $0.707$ & $0.680$ \\
    $w_{U1}{=}1.5$                & $33.3\%$ & $0.339$ & $0.552$ & $0.504$ & $0.700$ & $0.702$ \\
    $w_{U1}{=}2.0$ (released)     & $40.0\%$ & $0.500$ & $0.741$ & $0.485$ & $0.687$ & $0.727$ \\
    $w_{U1}{=}3.0$                & $50.0\%$ & $0.509$ & $0.756$ & $0.426$ & $0.497$ & $0.752$ \\
    \midrule
    \multicolumn{7}{l}{\emph{S1-2: U2-targeted distillation, U2--U4 weight $1.0$ each}} \\
    $w_{U1}{=}0.5$                & $14.3\%$ & $0.296$ & $0.561$ & $0.579$ & $0.730$ & $0.626$ \\
    $w_{U1}{=}1.0$ (released)     & $25.0\%$ & $0.334$ & $0.597$ & $0.568$ & $0.705$ & $0.641$ \\
    $w_{U1}{=}1.5$                & $33.3\%$ & $0.340$ & $0.631$ & $0.521$ & $0.630$ & $0.694$ \\
    $w_{U1}{=}2.0$                & $40.0\%$ & $0.369$ & $0.666$ & $0.478$ & $0.593$ & $0.713$ \\
    $w_{U1}{=}3.0$                & $50.0\%$ & $0.379$ & $0.714$ & $0.379$ & $0.563$ & $0.794$ \\
    \midrule
    \multicolumn{7}{l}{\emph{S2-1-1: understanding-tower replay mixture, $2$k steps}} \\
    no replay (S1-4)              & $0\%$    & $0.339$ & $0.599$ & $0.566$ & $0.706$ & $0.641$ \\
    $1{:}1$                       & $50.0\%$ & $0.376$ & $0.628$ & $0.560$ & $0.703$ & $0.668$ \\
    $2{:}1$ (released)            & $66.7\%$ & $0.374$ & $0.643$ & $0.561$ & $0.720$ & $0.650$ \\
    $3{:}1$                       & $75.0\%$ & $0.368$ & $0.647$ & $0.540$ & $0.720$ & $0.661$ \\
    $4{:}1$                       & $80.0\%$ & $0.339$ & $0.609$ & $0.505$ & $0.727$ & $0.728$ \\
    \bottomrule
  \end{tabular}
\end{table}

\begin{figure}[!p]
  \centering
  \includegraphics[width=0.72\textwidth]{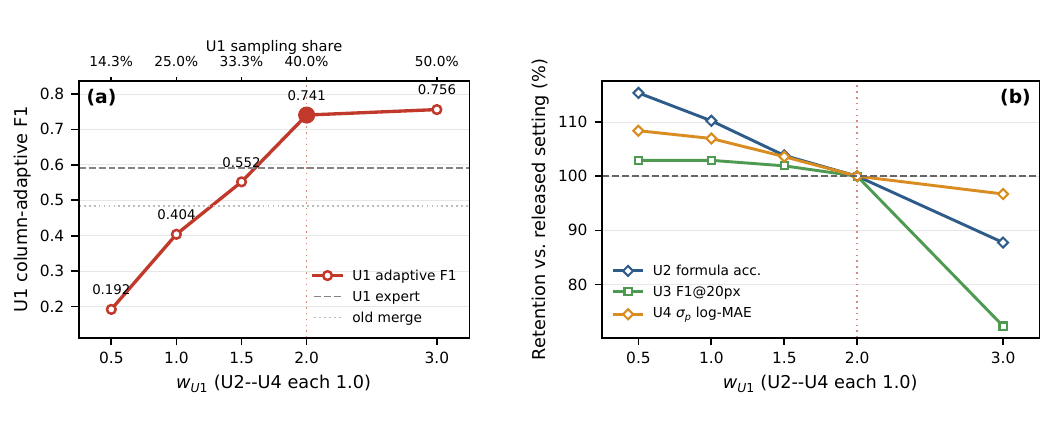}
  \caption{Effect of the U1 up-sampling weight in mixed understanding SFT (S1-1), with U2--U4 held at weight $1.0$. Panel~(a) shows U1 column-adaptive F1 ($k{=}0.5$), with the sampling share on the top axis. Panel~(b) shows retention of the other understanding metrics relative to the released $w_{U1}{=}2.0$ setting. The points are measured levels connected to guide the eye.}
  \label{fig:supp_ratio_s11}
\end{figure}

\begin{figure}[!p]
  \centering
  \includegraphics[width=0.72\textwidth]{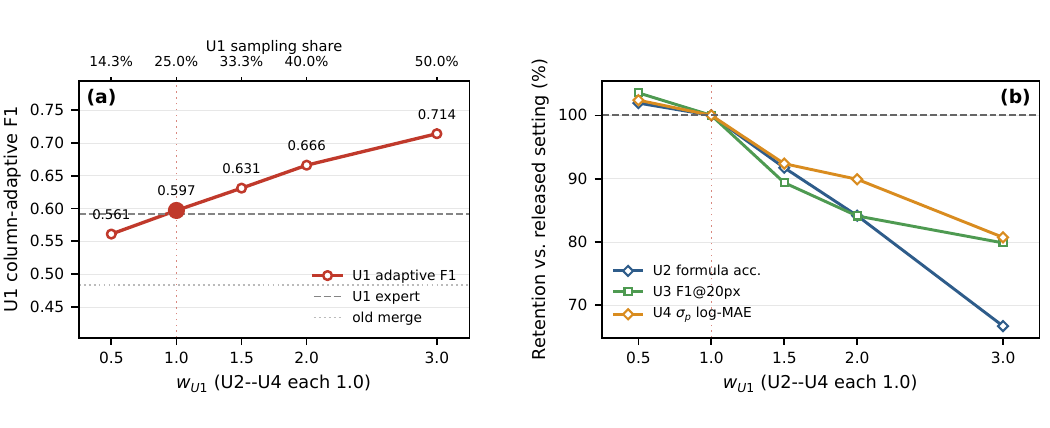}
  \caption{Effect of the U1 sampling weight during S1-2, starting from the released S1-1 checkpoint. Every level uses the same $5$k-step schedule and U2-targeted distillation, with U2--U4 at weight $1.0$. U1 is measured by column-adaptive F1 ($k{=}0.5$). Panel~(b) normalizes the other understanding metrics to the released $w_{U1}{=}1.0$ setting.}
  \label{fig:supp_ratio_s12}
\end{figure}

\begin{figure}[!p]
  \centering
  \includegraphics[width=0.72\textwidth]{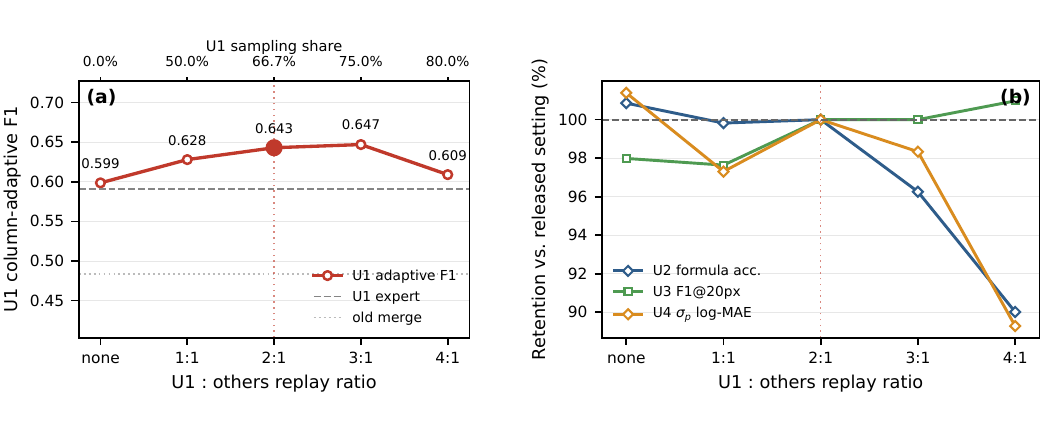}
  \caption{Effect of the S2-1-1 replay mixture, followed by the same S2-2 assembly. The leftmost point is the S1-4 checkpoint without replay. The other points are $2$k-step replay runs. Panel~(a) shows U1 column-adaptive F1 ($k{=}0.5$), and panel~(b) shows retention of U2--U4 relative to the released $2{:}1$ ratio. Generation scores are omitted because the assembled replay variants share the same generation tower from S2-1-2.}
  \label{fig:supp_ratio_replay}
\end{figure}

To explain the fall and recovery of U1 in \cref{fig:stage_progression}, we vary its sampling ratio at S1-1, S1-2, and S2-1-1. S1-1 starts from S0 and runs $10$k mixed-understanding SFT steps at learning rate $2\mathrm{e}{-5}$. S1-2 starts from the released S1-1 checkpoint and runs $5$k steps at $1\mathrm{e}{-5}$, with identical U2-targeted distillation at every tested ratio. In both stages, U2--U4 retain weight $1.0$ each and U1 receives a batch fraction $w_{U1}/(w_{U1}{+}3)$. The S2-1-1 sweep starts from S1-4, varies the U1-to-other-task replay ratio over $2$k cross-entropy steps at $1\mathrm{e}{-5}$, and uses the same S2-2 assembly. U1's replay fraction is $w_{U1}/(w_{U1}{+}1)$. \Cref{fig:supp_ratio_s11,fig:supp_ratio_s12,fig:supp_ratio_replay,tab:supp_ratio} reports full-test-set results, with U1 measured by the column-adaptive F1 used in \cref{fig:stage_progression}.

At S1-1, raising $w_{U1}$ from $0.5$ to the released $2.0$ increases U1 adaptive F1 from $0.192$ to $0.741$, a relative gain of $286\%$. Raising it further to $3.0$ improves U1 by only another $2.0\%$, while U2 formula accuracy drops by $12.2\%$, U3 F1@$20$\,px drops by $27.7\%$, and U4 log-MAE rises by $3.4\%$ relative to $w_{U1}{=}2.0$. The released weight therefore captures most of the U1 gain without the larger losses in other understanding tasks.

From S1-1 to S1-2, U1's sampling share falls from $40\%$ to $25\%$, a decrease of 15 percentage points, and its adaptive F1 falls by $19.4\%$ from $0.741$ to $0.597$. Within S1-2, restoring $w_{U1}$ from the released $1.0$ to $2.0$ raises U1 F1 by $11.6\%$ to $0.666$, recovering about $48\%$ of the stage-to-stage F1 drop. This U1 gain has a substantial trade-off. U2 formula accuracy and U3 F1@$20$\,px each fall by about $15.9\%$, while U4 log-MAE rises by $11.2\%$ relative to the released S1-2 ratio. The released weight preserves stronger results on U2--U4 while keeping U1 close to its expert. These controlled sweeps show that sampling allocation materially influences U1's decline and the balance among understanding tasks, complementing the gradient-alignment evidence in \cref{sec:supp_grad_probe}.

S2-1-1 replay raises U1 adaptive F1 from $0.599$ without replay to $0.643$ at the released $2{:}1$ ratio, a relative gain of $7.3\%$. Raising the ratio to $3{:}1$ improves U1 by only another $0.6\%$, while U2 formula accuracy falls by $3.7\%$ and U4 log-MAE rises by $1.7\%$. At $4{:}1$, U1 F1 itself falls by $5.3\%$ relative to $2{:}1$. The $2{:}1$ setting recovers U1 while retaining a more balanced result across the understanding tasks. Together, the three sweeps show that giving U1 more data improves its performance up to a point, but excessive emphasis substantially sacrifices U2--U4 performance.

\clearpage

